\documentclass[final]{IEEEtran}

\usepackage{graphicx,amsmath,amsfonts,amssymb,bm,hyperref,url,breakurl,epsfig,epsf,color,fullpage,MnSymbol,mathbbol,fmtcount,semtrans,cite,caption,subcaption,multirow,comment}
\usepackage[]{algorithm2e}
\usepackage[noend]{algpseudocode}

\usepackage[utf8]{inputenc} 
\usepackage[T1]{fontenc}    
\usepackage{hyperref}       
\usepackage{booktabs}       
\usepackage{amsfonts}       
\usepackage{nicefrac}       
\usepackage{microtype}      
\microtypesetup{expansion=false}
 
\makeatletter
\def\BState{\State\hskip-\ALG@thistlm}
\makeatother

\usepackage{mathtools}

\usepackage{tikz}
\usepackage{pgfplots}
\usetikzlibrary{pgfplots.groupplots}

\usepackage{movie15}

\usepackage[bottom,hang,flushmargin]{footmisc}

\usepackage{hyperref}
\definecolor{darkred}{RGB}{150,0,0}
\definecolor{darkgreen}{RGB}{0,150,0}
\definecolor{darkblue}{RGB}{0,0,200}
\hypersetup{colorlinks=true, linkcolor=darkred, citecolor=darkgreen, urlcolor=darkblue}

\newtheorem{theorem}{Theorem}[section]

\newtheorem{lemma}[theorem]{Lemma}

\newtheorem{definition}[theorem]{Definition}

\newcommand{\leqsym}[1]{\stackrel{\text{(#1)}}{\leq}}

\newcommand{\beq}{\begin{equation}}
\newcommand{\eeq}{\end{equation}}

\newcommand{\nn}{\nonumber}
\newcommand{\la}{\lambda}

\newcommand{\Lc}{{\cal{L}}}

\newcommand{\bteta}{\boldsymbol{\theta}}

\newcommand{\La}{{\boldsymbol{{\Lambda}}}}

\newcommand{\onebb}{{\mathbf{1}}}
\newcommand{\Iden}{{\mtx{I}}}
\newcommand{\M}{{\mtx{M}}}

\newcommand{\order}[1]{{\cal{O}}(#1)}

\newcommand{\z}{{\vct{z}}}

\newcommand{\tn}[1]{\|{#1}\|_{\ell_2}}
\newcommand{\tnn}[1]{\bigg\|{#1}\bigg\|_{\ell_2}}

\newcommand{\te}[1]{\|{#1}\|_{\psi_1}}
\newcommand{\tsub}[1]{\|{#1}\|_{\psi_2}}

\newcommand{\Cc}{\mathcal{C}}
\newcommand{\Cext}{\mathcal{Cc}_{\text{ext}}}

\newcommand{\Kc}{\mathcal{K}}
\newcommand{\Rc}{\mathcal{R}}

\newcommand{\bbeta}{{\boldsymbol{\beta}}}

\newcommand{\Bc}{\mathcal{B}}
\newcommand{\Sc}{\mathcal{S}}

\newcommand{\Nn}{\mathcal{N}}

\newcommand{\vb}{\vct{v}}

\newcommand{\w}{\vct{w}}

\newcommand{\li}{\left<}
\newcommand{\ri}{\right>}

\newcommand{\ab}{\vct{a}}
\newcommand{\bb}{\vct{b}}
\newcommand{\ub}{{\vct{u}}}

\newcommand{\h}{\vct{h}}

\newcommand{\g}{{\vct{g}}}

\newcommand{\Tc}{\mathcal{T}}

\newcommand{\mat}[1]{{\text{mat}{#1}}}

\newcommand{\wtail}{w_{\text{tail}}}
\newcommand{\xt}{\tilde{\x}}

\newcommand{\x}{\vct{x}}

\newcommand{\y}{\vct{y}}

\newcommand{\bgl}{{~\big |~}}

\definecolor{emmanuel}{RGB}{255,127,0}

\newcommand{\clconv}{{{\text{${{\bf{\text{clconv}}}}$}}}}
\newcommand{\close}{{{\text{${{\bf{\text{cl}}}}$}}}}

\newcommand{\R}{\mathbb{R}}
\newcommand{\Pro}{\mathbb{P}}

\newcommand{\sgn}[1]{\textrm{sgn}(#1)}
\renewcommand{\P}{\operatorname{\mathbb{P}}}
\newcommand{\E}{\operatorname{\mathbb{E}}}

\newcommand{\e}{\vct{e}}

\newcommand{\vct}[1]{\bm{#1}}
\newcommand{\mtx}[1]{\bm{#1}}

\newcommand{\ut}{\tilde{\vct{u}}}
\newcommand{\vt}{\tilde{\vct{v}}}
\newcommand{\co}{\mathbf{1}}

\newcommand{\rank}{\operatorname{rank}}

\newcommand{\Pc}{{\cal{P}}}
\newcommand{\X}{{\mtx{X}}}

\graphicspath{{Authors/}}
\begin{document}
\title{Quickly Finding the Best Linear Model in High Dimensions}
\author{Yahya~Sattar\quad\quad  \quad \quad Samet~Oymak
\thanks{Department of Electrical and Computer Engineering, University of California, Riverside, CA 92521, USA. Email: ysatt001@ucr.edu, oymak@ece.ucr.edu.}
}


\maketitle
\begin{abstract}
We study the problem of finding the best linear model that can minimize least-squares loss given a dataset. While this problem is trivial in the low-dimensional regime, it becomes more interesting in high-dimensions where the population minimizer is assumed to lie on a manifold such as sparse vectors. We propose projected gradient descent~(PGD) algorithm to estimate the  population minimizer in the finite sample regime. We establish linear convergence rate and data-dependent estimation error bounds for PGD. Our contributions include: 1) The results are established for heavier tailed sub-exponential distributions besides sub-gaussian. 2) We directly analyze the empirical risk minimization and do not require a realizable model that connects input data and labels. 3) Our PGD algorithm is augmented to learn the bias terms which boosts the performance. The numerical experiments validate our theoretical results.
\end{abstract}
\IEEEpeerreviewmaketitle
\vspace{-7pt}\section{Introduction}\label{sec:Intro}\vspace{-4pt}
Supervised learning is concerned with finding a relation between the input-output pairs $(\x_i,y_i)_{i=1}^n\in\R^{p}\times \R $. The simplest relations are linear functions where the output $y_i$ is estimated by a linear function of the input, that is, $\hat{y}_i=\li \x_i, \bteta \ri$. Using quadratic loss, we can find the optimal $\bteta$ with a simple linear regression which minimizes.\vspace{-5pt}
\[
\Lc(\bteta)=\frac{1}{2}\sum_{i=1}^n(y_i-\li\bteta,\x_i\ri)^2.\vspace{-5pt}
\]
If the samples are i.i.d.~and input has identity covariance, the population minimizer ($n\rightarrow\infty$) is simply given by 
\[
\bteta^\star=\arg\min_{\bteta} \E[\Lc(\bteta)]=\E[y\x].
\]
where $(\x,y)$ is drawn from same distribution as data.  In many applications, we operate in the high-dimensional regime where we have fewer samples than the parameter dimension i.e.~$n\ll p$. In this case, the problem is ill-posed; however, if $\bteta^\star$ lies on a low-dimensional manifold, we can take advantage of this information to solve the problem. We assume $\bteta^\star$ is structured-sparse, for instance, it can be a signal that is sparse in a dictionary or it can be a low-rank matrix. If $\Rc$ is a regularization function that promotes this structure, we can solve the regularized empirical risk minimization (ERM)
\begin{align}
\hat{\bteta}=\arg\min_{\bteta}\frac{1}{2}\tn{\y-\X\bteta}^2~~\text{subject to}~~\Rc(\bteta)\leq R.\label{PC}
\end{align}
where $\y=[y_1~\dots~y_n]^T\in\R^{n}~\textrm{and}~\X=[\x_1~\dots~\x_n]^T\in\R^{n\times p}$ are the output labels and data matrix respectively. This problem is well-studied in the statistics and compressed sensing (CS) literature. However, much of the theory literature is concerned with the scenario where the problem is realizable i.e. the outputs are explicitly generated with respect to some ground truth vector $\ab$. In the simplest scenario, input/output relation can be $y= \li \x, \ab \ri+\z$ where $\z$ is independent zero-mean noise vector. In this case, one simply has $\bteta^\star=\ab$. Such realizability assumption is also common in the single-index models \cite{plan2016high,boufounos20081}. One contribution of this paper will be analyzing regularized ERM without the realizability assumption. 


Bias in the data can negatively affect the estimation quality. Assuming input is zero-mean, instead of solving \eqref{PC} we can solve a modified problem which accounts for the mean of the output as well. Again, denoting the regularization function by $\Rc$, we will solve the modified problem 
\begin{align}
\hat{\bteta},\hat{\mu}=\arg\min_{\bteta,\mu}\Lc(\bteta,\mu)~\text{subject to}~\Rc(\bteta)\leq R.\label{PC_modified}
\end{align} 
where the loss is given by $\Lc(\bteta,\mu)=\frac{1}{2}\big\|\y-[\X~\co]\begin{bmatrix}\bteta\\\mu\end{bmatrix}\big\|_{\ell_2}^2$. We will show that solving problem~\eqref{PC_modified} is essentially equivalent to solving~\eqref{PC} with debiased output hence it will result in more accurate estimation. The goal of this paper is studying problem \eqref{PC_modified} under a general algorithmic framework, establishing finite-sample statistical and algorithmic convergence, and addressing practical considerations on the data distribution. In particular, we are interested in how well one can estimate the best linear model (BLM) given by the pair $(\bteta^\star=\E[y\x],\mu^\star=\E[y])$. For estimation, we will utilize the projected gradient descent algorithm given by the iterates 
\begin{equation}\label{PGD_algo}
\begin{aligned}
&{\bteta}_{\tau+1}=\Pc_{\Kc}({\bteta}_{\tau}-\eta\nabla\Lc_{\bteta}({{\bteta}_{\tau},{\mu}_{\tau}})),\\
&{\mu}_{\tau+1}={\mu}_{\tau}-\eta\nabla\Lc_{\mu}({{\bteta}_{\tau},{\mu}_{\tau}}),
\end{aligned}
\end{equation}
where $\Pc_\Kc$ projects onto the constraint set $\Kc=\{\bteta\in \R^{p} \bgl \mathcal{R}(\bteta) \leq R\}$ and $\eta$ is the step size.
\subsection{Relation to Prior Work}
There is a significant amount of literature on nonlinear (or one-bit) CS \cite{ganti2015learning,oymak2017fast,plan2017high,plan2016generalized,thrampoulidis2015lasso,boufounos20081,jacques2013robust,vershynin2015estimation,dirksen2017one,dirksen2018robust_circ,plan2013robust}. \cite{agarwal2010fast,oymak2017fast,oymak2015sharp,giryes2018tradeoffs,beck2009fast} study algorithmic and statistical convergence rates for first order methods such as projected/proximal gradient descent. For nonlinear CS, \cite{genzel2017high,oymak2017fast,thrampoulidis2015lasso,plan2017high} provide statistical analysis of single index estimation with a focus on Gaussian data. Recently, one-bit CS techniques have been extended to sub-gaussian distributions using dithering trick which adds noise before quantization \cite{dirksen2018robust,thrampoulidis2018generalized,jacques2017time,xu2018quantized}. Dithering is introduced to guarantee consistent estimation of the ground-truth parameter. The papers~\cite{yang2016sparse,yang2017high,yang2017learning,yang2017stein,yap2013stable} address non-gaussianity by utilizing Stein identity which requires access to the distribution of the input samples. Closer to us \cite{genzel2018mismatch} studies the constrained empirical risk minimization with linear functions and squared loss with a focus on convex problems. In comparison our analysis applies to a broader class of distributions and focus on first order algorithms. Much of our analysis focuses on addressing subexponential samples, which requires tools from high-dimensional probability \cite{talagrand2014gaussian,oymak2018learning}. 

Our results apply to general regularizers and borrow ideas from \cite{oymak2017fast,plan2017high,plan2016generalized,thrampoulidis2015lasso}. Similar to these, we view the nonlinearity between input and output as an additive noise. The convergence analysis of projected gradient descent is a rather well-understood topic and we utilize insights from \cite{giryes2018tradeoffs,oymak2015sharp,agarwal2010fast,beck2009fast} for our analysis.

\subsection{Contributions}
At a high-level our work has three distinguishing features compared to the prior literature.

\vspace{-2pt}
\noindent~~~$\bullet$ {Subexponential samples:} Most nonlinear CS results apply to Gaussian or subgaussian data when dithering trick is utilized \cite{dirksen2018robust,thrampoulidis2018generalized,jacques2017time,xu2018quantized}. We take advantage of the recent techniques for subexponential distributions to provide statistical/computational guarantees for heavier-tailed distributions.

\vspace{-2pt}
\noindent~~~$\bullet$ {No realizability assumption:} Nonlinear CS literature is typically concerned with a ground-truth vector to be recovered. For instance, one-bit CS aims to learn $\bteta$ from samples of type $y=\sgn{\bteta^T\x}$. Unlike these, we do not enforce such relationship to exist between input and output, hence the results apply under much weaker assumptions. Instead of a ground-truth $\bteta$, we work with the population BLM $\bteta^\star$. However, $\bteta^\star$ can be shown to coincide with ground truth when it exists, if the input distribution is {\em{nice}} (e.g.~Gaussian) \cite{oymak2017fast,plan2017high,plan2016generalized,thrampoulidis2015lasso}.



\vspace{-2pt}
\noindent$~~~\bullet$ {Bias estimation:} Our analysis addresses the bias in the output by solving the modified problem~\eqref{PC_modified}. We show that \eqref{PC_modified} can be studied in a similar fashion to \eqref{PC} by studying the statistical properties of the concatenated data matrix. However, empirically this modification results in a substantial improvement in estimation.
\subsection{Paper Organization}
We review mathematical background and formulate the problem in Section \ref{sec:math_prelim}. 
We introduce our main results on statistical and computational convergence guarantees in Section \ref{sec: main_reslts}. Section \ref{sec: exp} provides numerical experiments to corroborate our theoretical results. Proofs of the main results are provided in Section \ref{sec: proof} and finally the concluding remarks
are made in Section \ref{sec: conclusion}.
\section{Preliminaries and Problem Formulation}\label{sec:math_prelim}
In this section we introduce statistical quantities which are utilized to characterize the benefits of the regularization $\Rc$.

We first set the notation.  $c,c_0,\dots, C$ denote positive absolute constants. For a vector $\vb$, we denote its Euclidean norm by $\tn{\vb}$ and its $\ell_{\infty}$ norm by $\|\vb\|_{\infty}$. Similarly for a matrix $\X$, we denote its spectral norm by $\|\X\|$. Given a set $S$, let $\close(S)$ and $\clconv(S)$ be the minimal closed set and minimal closed-convex set containing $S$ respectively. Let $\text{rad}(S)$ denote the set radius $\sup_{\vb\in S}\tn{\vb}$. For closed sets, let $\Pc_{S}(\cdot)$ be the projection operator defined as $\Pc_{S}(\ab)=\arg\min_{\vb\in S}\tn{\ab-\vb}$. $\Nn(\mu,\sigma^2)$ denotes the normal distribution and $\Bc^{p}$ denote the unit ball in $\R^p$. $\onebb$ is the all ones vector of proper dimension. We will use $\gtrsim$ and $\lesssim$ for inequalities that hold up to a constant factor.


Suppose we are given $n$ i.i.d.~samples $(\x_i,y_i)_{i=1}^n \sim (\x,y)$. To keep the exposition clean, we assume that $\x$ is whitened, that is, it has zero-mean and identity covariance. We will aim to find a linear relation between the modified input-output pairs $([\x_i^T~1]^T,y_i)_{i=1}^n$. Let us consider the statistical properties of our modified estimate in the population limit which is given by 
\begin{align}
 \bteta^\star , \mu^\star &=\arg\min_{\bteta,\mu} \E[\Lc(\bteta,\mu)] \nn \\ 
&=\E[y\x],\E[y].\nn
\end{align}
Thus, in the limiting case, $\mu^\star$ captures the mean of the output and $\bteta^\star$ is the ideal solution of the problem with debiased output. Our goal is estimating the population minimizer ${\bteta^\star},\mu^\star$; which minimizes the expected quadratic loss $\E[(y-\bteta^T\x-\mu)^2]$. As discussed in Section~\ref{sec:Intro}, assuming $\bteta^\star$ is structured sparse, we consider a non-asymptotic estimation of ${\bteta^\star},\mu^\star$ via problem~\eqref{PC_modified}. 
To proceed with analysis, set 
\begin{align}
\mathcal{K} &= \{\bteta\in \R^{p} \bgl \mathcal{R}(\bteta) \leq R\}, \label{def:K sets} \\
\mathcal{K}_{\text{ext}} &= \{[\bteta^T~\mu]^T\in \R^{p+1} \bgl \mathcal{R}(\bteta) \leq R\}.\label{def:Kext sets}
\end{align}
We investigate the PGD algorithm \eqref{PGD_algo} which can be written as
\begin{equation}\label{eqn:PGD_update}
\hspace{-6pt}\begin{bmatrix} \bteta_{\tau+1} \\ \mu_{\tau+1} \end{bmatrix} = \mathcal{P}_{\mathcal{K}_{\text{ext}}}\bigg(\begin{bmatrix} \bteta_{\tau} \\ \mu_{\tau} \end{bmatrix} + \eta[\X~\co]^T\bigg(\bm{y}-[\X~\co]\begin{bmatrix} \bteta_{\tau} \\ \mu_{\tau} \end{bmatrix}\bigg)\bigg),
\end{equation}
where $\eta$ is a fixed learning rate and $[\X~\co]\in\R^{n\times (p+1)}$ is the modified data matrix constructed as follows
\[
[\X~\co]=\begin{bmatrix}\x_1^T~1\\\vdots\\\x_n^T~1\end{bmatrix}.
\]
Following \cite{oymak2017fast,chandrasekaran2012convex} PGD analysis can be related to the {\em{tangent ball}} around the population parameter $\bteta^\star$ which is given by
\begin{align}
\Cc=\close(\{\alpha \vb \bgl \vb+\bteta^\star\in\mathcal{K},~\alpha\geq 0 \})\bigcap \Bc^{p}.\label{def: tangent cone}
\end{align}
Similarly, we define the extended tangent ball as follows
	\begin{align}
	\Cext=\bigg\{ \begin{bmatrix} \alpha\vb \\ \gamma \end{bmatrix} \bgl \alpha\geq 0,~\vb\in\Cc,~\gamma\in\R \bigg\}\bigcap \Bc^{p+1}.\label{def: ext tangent cone}
	\end{align}
The two definitions above are closely related. For any vector $\vb \in \Cc$, we have that $[\sqrt{1-\gamma^2}\vb^T~\gamma]^T \in \Cext$ for $|\gamma|\leq 1$. In the following we will express the convergence rates and residual errors of the PGD algorithm \eqref{PGD_algo} in terms of the statistical properties of the tangent balls . 

\noindent {\bf{Technical approach:}} Denoting the parameter estimation error in \eqref{eqn:PGD_update} by $\h_{\tau}=[{\bteta_{\tau}}^T~\mu_{\tau}]^T-[{\bteta^\star}^T~\mu^\star]^T$ and the effective noise by $\w = \y-[\X~\co][{\bteta^\star}^T~\mu^\star]^T$, the PGD update can be shown to obey \cite{oymak2015sharp} (see Eq. (VI.10))
\begin{equation}\label{rec:main_recursion}
\tn{\h_{\tau+1}} \leq   \kappa \left( \tn{\h_{\tau}}\rho(\Cc) +\eta \nu(\Cc)\right)
\end{equation}
where $ \kappa$ is a numerical constant which is equal to $1$ for convex regularizer $\Rc$ and $2$ for arbitrary $\Rc$ and
\begin{align}
&\rho(\Cc)=\sup_{\ub,\vb \in \Cext}|\ub^T (\Iden-\eta[\X~\co]^T[\X~\co])\vb|,\label{eqn:conv+noise_form} \\
&\nu(\Cc)=\sup_{\vb \in \Cext} |\vb^T [\X~\co]^T \w|.\label{eqn:noise_form}
\end{align}
Here $\rho$ captures the algorithmic convergence and $\nu$ captures the statistical accuracy in terms of regularization. To achieve statistical learning bounds, we need to characterize the quantities above in finite sample. Existing literature provides a fairly good understanding of the related terms when $\X$ has subgaussian rows or $\w$ is independent of $\X$. The technical contributions of this work are i) extending these results to subexponential samples, ii) allowing for nonlinear dependencies between the noise and data, and iii) addressing the bias term by studying the concatenated matrix $[\X~\co]$. To proceed with statistical analysis, we introduce Gaussian width.
\begin{definition}[(Perturbed) Gaussian width~\cite{oymak2018learning}] \label{def:pert_gauss} The Gaussian width of a set $S \subset \Bc^p$ is defined as 
\[
\omega(S)= \E_{\g\sim\Nn(0,\Iden_p)}[\sup_{\vb\in S} \vb^T\g].
\] Let $C>0$ be an absolute constant. Given an integer $n\geq 1$, the perturbed Gaussian width $\omega_n(T)$ of $T\subset\Bc^d$ is defined as
	\[
	\omega_n(T)=\underset{\text{rad}(S)\leq C}{\min_{\clconv(S)\supseteq T}}\omega(S)+\frac{\gamma_1(S)}{\sqrt{n}}
	\]
	where $\gamma_1(S)$ is Talagrand's $\gamma_1$-functional (see~\cite{talagrand2014gaussian}) with $\ell_2$-metric.
\end{definition}
Gaussian width helps to quantify the complexity of the regularized problem and determines the sample complexity of the linear inverse problems i.e.~high-dimensional problems become manageable in the regime $n\gtrsim \omega^2(\Cc)$ \cite{McCoy,chandrasekaran2012convex}. Perturbed width is introduced more recently in \cite{oymak2018learning} to address subexponential samples. \cite{oymak2018learning} shows that, for standard regularizers such as $\ell_0,\ell_1$, subspace, and rank constraints, we have that 
\begin{align}
\omega^2(\Cc)\sim \omega_n^2(\Cc)\label{near equal}
\end{align} in the interesting regime $n\geq \omega^2(\Cc)$. Hence, perturbed width has the same statistical accuracy of Gaussian width but applies to subexponential samples.

As illustrated in Table \ref{table:gauss width}, square of the Gaussian width captures the degrees of freedom for practical regularizers. Table \ref{table:gauss width} is obtained by setting $R = \Rc(\bteta^\star)$ in \eqref{def:K sets}. In practice, a good choice for $R$ can be found by using cross validation. It
is also known that the performance of PGD is robust to choice of $R$ (see Thm 2.6 of \cite{oymak2015sharp}).
\begin{center}
	{
		\setlength\arrayrulewidth{1pt}
		\resizebox{\columnwidth}{!}{	
			\begin{tabular}{ |c|c|c|c| }
				\hline
				\textbf{Constraint}& \textbf{Parameter vector model} & $\boldsymbol{\omega^2(\Cc)}$ \\ 
				\hline
				None & $\bteta^\star\in\R^p$ & ${p}$ \\ 
				\hline
				Sparsity $\|\cdot\|_{\ell_0}$ & $s$ non-zero entries & ${s\log(6p/s)}$ \\ 
				\hline
				$\ell_1$ norm $\|\cdot\|_{\ell_1}$ & $s$ non-zero entries & ${s\log(6p/s)}$ \\ 
				\hline
				Subspace & $\bteta^\star\in \Sc, \; \dim(\Sc)=k$ & ${k}$ \\ 
				\hline
				Matrix rank & $\rank(\mat{(\bteta^\star)})\leq r$ & ${rp^{1/2}}$ \\ 
				\hline
			\end{tabular}
		}
	}

	\captionof{table}{List of low-dimensional models and corresponding Gaussian widths (up to a constant factor) for the constraint sets $\Kc = \{\bteta \bgl \Rc(\bteta)\leq \Rc(\bteta^\star)\}$. If constraint is set membership such as subspace, $\Rc(\bteta)=0$ inside the set and $\infty$ outside. Furthermore, we represent the vector $\bteta^\star\in \R^p$ in matrix form as $\mat{(\bteta^\star)} \in \R^{\sqrt{p}\times \sqrt{p}}$.}	\label{table:gauss width}
\end{center}
The next statistical quantity required in our analysis is the Orlicz norm defined as.
\begin{definition}[Orlicz norms] \label{def:orlicz_norm}For a scalar random variable Orlicz-$a$ norm is defined as
	\[
	\|X\|_{\psi_{a}}=\sup_{p\geq 1}p^{-1/a}(\E[|X|^p])^{1/p}
	\]
	Orlicz-$a$ norm of a vector $\x\in\R^d$ is defined as $\|\x\|_{\psi_{a}}=\sup_{\vb\in \Bc^{d}} \|\vb^T\x\|_{\psi_{a}}$. Subexponential and subgaussian norms are special cases of Orlicz-$a$ norm given by $\te{\cdot}$ and $\tsub{\cdot}$ respectively.
\end{definition}
Based on perturbed Gaussian width definition, we will show that one can upper bound the critical quantities \eqref{eqn:conv+noise_form} and \eqref{eqn:noise_form}. In return, this will reveal the statistical and computational performance of the PGD algorithm. This is the topic of the next section which states our main results.



\section{Main Results}\label{sec: main_reslts}
In this section we estimate the convergence rate and the statistical accuracy of the PGD algorithm as a function of sample size, complexity of the parameter (e.g.~sparsity level), and the distribution of the data (whether subgaussian or subexponential). Our main theorem establishes a linear convergence rate of PGD and shows that PGD achieves statistically efficient error rates. We first describe the data model.
\begin{definition}[Isotropic vector] $\x\in\R^p$ is called an isotropic Orlicz-$a$ vector if it is zero-mean with identity covariance and if its Orlicz-$a$ norm $\|\x\|_{\psi_{a}}$ is bounded by an absolute constant.
\end{definition}
\begin{definition} [$\sigma$-noisy datasets] We assume the samples $(y_i,\x_i)_{i=1}^n$. We call a dataset $\sigma$-Orlicz-$a$ if the input samples are isotropic Orlicz-a vectors and the residual at the ground truth obeys
\[
\|y-\x^T\bteta^\star-\mu^\star\|_{\psi_{a}}\leq \sigma.
\]
We call $\sigma$-Orlicz-$1$ dataset $\sigma$-subexponential and $\sigma$-Orlicz-$2$ dataset $\sigma$-subgaussian.
\end{definition}
Note that residual at the ground truth is the noise in our problem which may be function of the nonlinearity. Our main results capture the PGD performance for different dataset models.
\begin{theorem}[Subgaussian]\label{thrm:subgaussian_convergence}
	Suppose $(\x_i,y_i)_{i=1}^n$ is a $\sigma$-subgaussian dataset. Assume $n \gtrsim {(\omega(\Cc)+t)^2}$ and set learning rate $\eta=1/n$. Let $\Rc$ be an arbitrary regularizer. Starting form any initial estimate $[\bteta_0^T~\mu_0]^T$, with probability at least $1-6\exp(-c_0 t^2/2)-4n^{-100}$, all PGD iterates \eqref{eqn:PGD_update}~obeys
	\begin{equation}
	\begin{aligned}
	\tnn{\begin{bmatrix} \bteta_{\tau}-\bteta^\star \\ \mu_{\tau}-\mu^\star \end{bmatrix}} \leq~& (c\frac{{\omega(\Cc)+t}}{\sqrt{n}})^{\tau}\tnn{\begin{bmatrix} \bteta_{0}-\bteta^\star \\ \mu_{0}-\mu^\star \end{bmatrix}}\\
	&+C\sigma  {\frac{(\omega(\Cc)+t)\sqrt{\log(n)}}{\sqrt{n}}}.\nonumber
	\end{aligned}
	\end{equation}
\end{theorem}
Similarly, for subexponential samples, we have the following theorem which applies to convex regularizers.
\begin{theorem}[Subexponential]\label{thrm:subexp_convergence}
	Suppose $(\x_i,y_i)_{i=1}^n$ is a $\sigma$-subexponential dataset. Set $q=(n+p)\log^3(n+p)$. Set learning rate $\eta={c_0/q}$, suppose $\Rc$ is convex and $n \gtrsim {(\omega_n(\Cc)+t)^2}$. Starting from initialization $[\bteta_0^T~\mu_{0}]^T$, with probability at least $1-9\exp(-c_0{\min(n,t\sqrt{n},t^2)})-5(n+p)^{-100}$, all PGD iterates \eqref{eqn:PGD_update} obey
	\begin{equation}
	\begin{aligned}
	\tnn{\begin{bmatrix} \bteta_{\tau}-\bteta^\star \\ \mu_{\tau}-\mu^\star \end{bmatrix}} &\leq \left(1-{\frac{cn}{q}}\right)^{\tau}\tnn{\begin{bmatrix} \bteta_{0}-\bteta^\star \\ \mu_{0}-\mu^\star \end{bmatrix}}\\
	&+C\sigma \frac{(\omega_n(\Cc)+t)\log(n)}{\sqrt{n}}.\nonumber
	\end{aligned}
	\end{equation}
\end{theorem}

Both of these results show that PGD iterates converge to population parameters $\bteta^\star,~\mu^\star$ at a linear rate. Subexponential theorem requires a more conservative choice of learning rate. The statistical estimation error grows as $\omega(\Cc)/\sqrt{n}$ for subgaussian and $\omega_n(\Cc)/\sqrt{n}$ for subexponential. Since our results apply in the regime $n\gtrsim \omega^2(\Cc)$, following \eqref{near equal}, statistical errors associated with subgaussian and subexponential are same up to a constant for typical regularizers.

Our main results follow from Theorems~\ref{thrm:convergence_rates} and \ref{thrm:error_bound} which are the topics of the following sections.

\subsection{Controlling the Convergence Rate of PGD}
In this section, we study the convergence rate characterized by the $\rho(\Cc)$ term. The challenges we address are (i) characterizing the {\em{restricted singular values}} of the subexponential data matrices and (ii) addressing the concatenated all ones vector.
\begin{theorem}[Convergence rate]\label{thrm:convergence_rates}
	Suppose $(\x_i,y_i)_{i=1}^n$ is a $\sigma$-subgaussian dataset and $[\X~\co]$ is the modified-data matrix, where $\co$ is a vector of all ones. Let $\Cc$ and $\Cext$ be the tangent balls as defined in \eqref{def: tangent cone} and \eqref{def: ext tangent cone} respectively. Assume $n\gtrsim {(\omega(\Cc)+t)^2}$. Setting $\eta=1/n$, with probability at least $1-4e^{-t^2}$ we have
	\[
	\rho(\Cc) \lesssim {\frac{\omega(\Cc)+t}{\sqrt{n}}}.
	\]
	If the dataset is $\sigma$-subexponential, then setting $\eta={c_0/(n+p)\log^3(n+p)}$ and assuming $n\gtrsim (\omega_n(\Cc)+t)^2$, with probability $1-5\exp(-c\min(n,t\sqrt{n},t^2))-3(n+p)^{-100}$, we have
	\begin{align*}
	\rho(\Cc) \leq 1-C_0\eta n.
	\end{align*}
\end{theorem}

Note that, subexponential requires a smaller choice of learning rate which results in slower convergence. 

\subsection{Bounding the Error due to Nonlinearity}
Next, we provide a bound on the effective noise level $\nu(\Cc)$; which is crucial for assessing statistical accuracy. This term arises from the nonlinearity and noise associated with the relation between input and output. For example, for single-index models, we have $\E[y\bgl\x]=\phi(\x^T\bteta_{\text{GT}})$ for some link function $\phi$ and ground truth $\bteta_{\text{GT}}$, and $\phi$ becomes the source of the nonlinearity. Our approach is similar to~\cite{oymak2017fast,plan2017high,genzel2018mismatch,plan2016generalized,thrampoulidis2015lasso} and treats the nonlinearity as a noise. The finite sample noise is captured by the residual vector 
\begin{align}
\w=\y-\X{\bteta^\star}-\co\mu^\star.\label{residual vector}
\end{align} Following $\nu(\Cc)$ term in \eqref{eqn:noise_form}, the contribution of the residual $\w$ to the estimated parameter is captured by the vector 
\begin{align}
\e=[\X~\co]^T\w=\sum_{i=1}^n(y_i-\mu^\star-\x_i^T\bteta^\star)\begin{bmatrix}\x_i\\ 1 \end{bmatrix}.
\end{align} Our key observation is that the properties of $\e$ can be characterized under fairly general assumptions compared to the existing literature; which is mostly restricted to zero-mean subgaussian samples.
 \begin{theorem}[Statistical error]\label{thrm:error_bound}
 	Suppose $(\x_i,y_i)_{i=1}^n\sim(\x,y)$ is a $\sigma$-subgaussian dataset. Let the tangent balls $\Cc$ and $\Cext$ be as defined in \eqref{def: tangent cone} and \eqref{def: ext tangent cone} respectively. Assume $n\gtrsim {(\omega(\Cc)+t)^2}$. Then, with probability at least $1-2\exp(-t^2/2)-4n^{-100}$, we have
 	\[
 	\frac{\nu(\Cc)}{n} \lesssim {\frac{\sigma(\omega(\Cc)+t)\sqrt{\log(n)}}{\sqrt{n}}}.
 	\]
 	where $\nu(\Cc)$ is the effective noise given by \eqref{eqn:noise_form}. If $(\x_i,y_i)_{i=1}^n$ is a $\sigma$-subexponential dataset and $n\gtrsim {(\omega_n(\Cc)+t)^2}$, with probability at least $1-4\exp(-c~{\min(t\sqrt{n},t^2)})-2n^{-100}$, we have
 	\[
 	\frac{\nu(\Cc)}{n}  \lesssim {\frac{\sigma(\omega_n(\Cc)+t)\log(n)}{\sqrt{n}}}.
 	\]  
 \end{theorem}
This theorem establishes the crucial finite sample upper bounds on $\nu(\Cc)$ for both subgaussian and subexponential data as a function of Gaussian width of the tangent ball. Combining our bounds on $\rho(\Cc)$ and $\nu(\Cc)$ and utilizing the recursion~\eqref{rec:main_recursion}, we can obtain the PGD convergence characteristics and prove the main theorems. 
\section{Numerical Experiments}\label{sec: exp}\vspace{-2pt}

\begin{figure}[t!]
\begin{centering}
\begin{subfigure}[t]{1.5in}
\includegraphics[height=0.8\linewidth,width=1\linewidth]{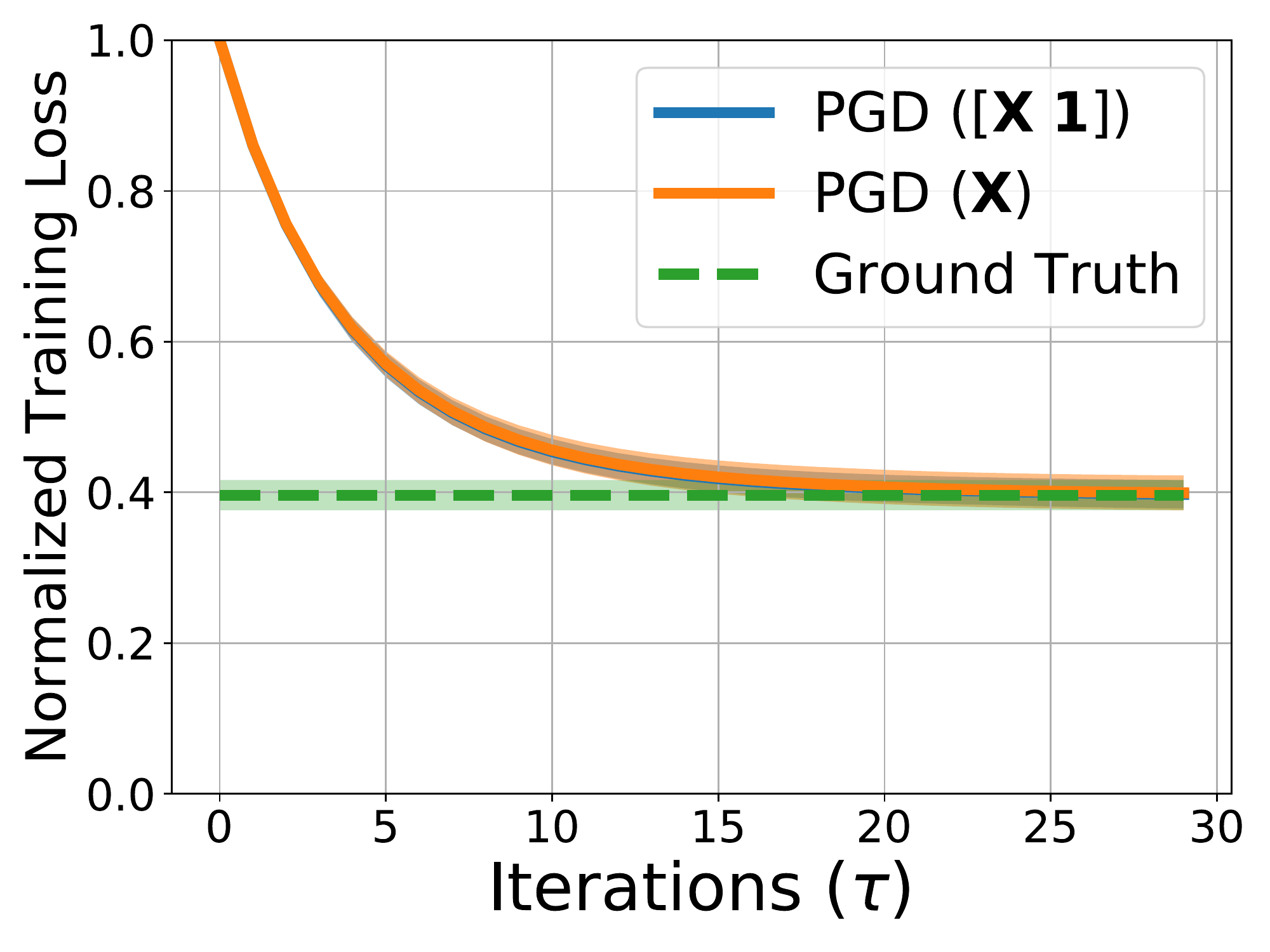}\vspace{-5pt}\subcaption{$\phi=\text{sign}$}\label{fig1a}
\end{subfigure}
\end{centering}~
\begin{centering}
\begin{subfigure}[t]{1.5in}
\includegraphics[height=0.8\linewidth,width=1\linewidth]{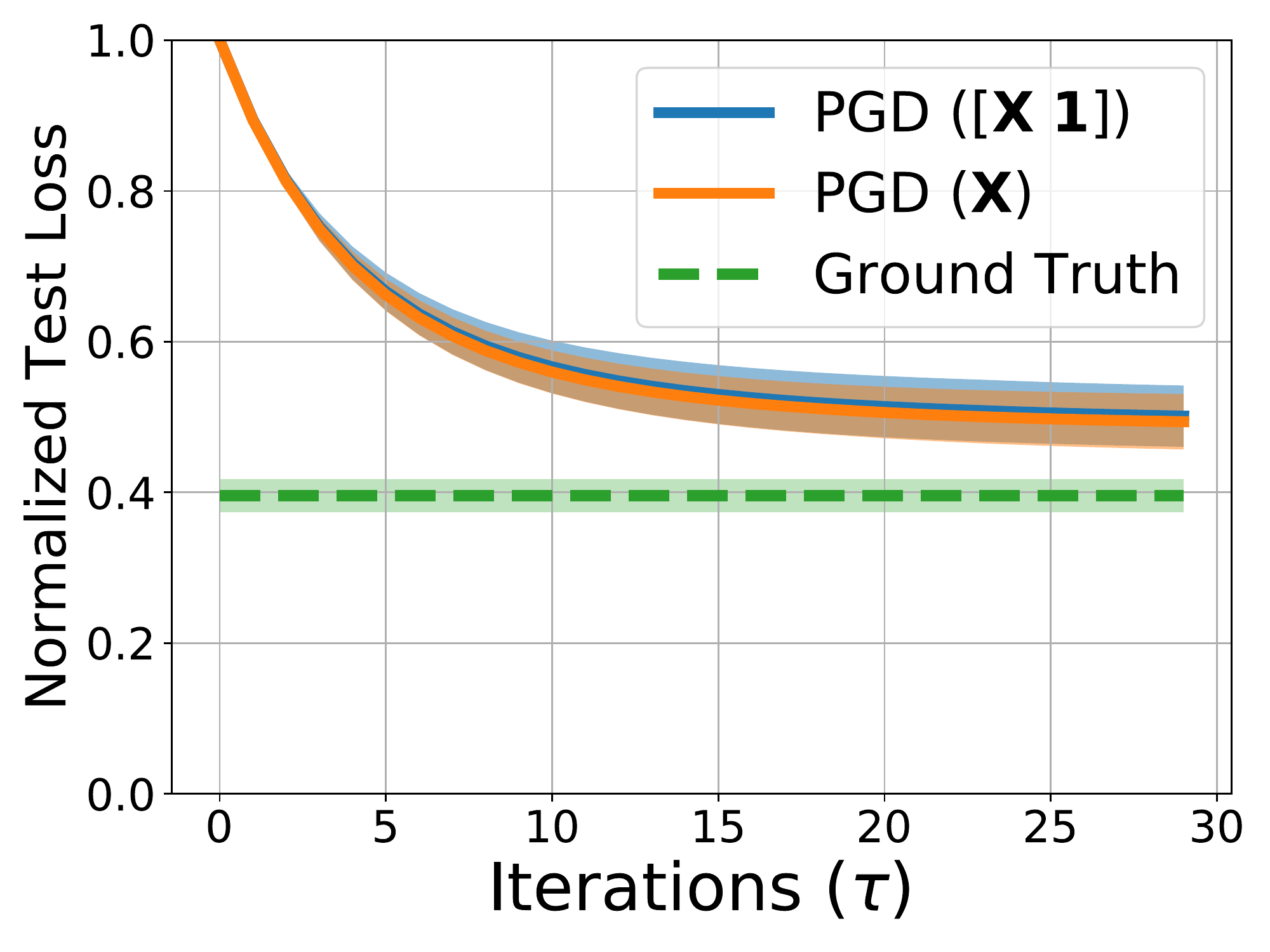}\vspace{-5pt}\subcaption{$\phi=\text{sign}$}\label{fig1b}
\end{subfigure}
\end{centering}\\
\begin{centering}
\begin{subfigure}[t]{1.5in}
\includegraphics[height=0.8\linewidth,width=1\linewidth]{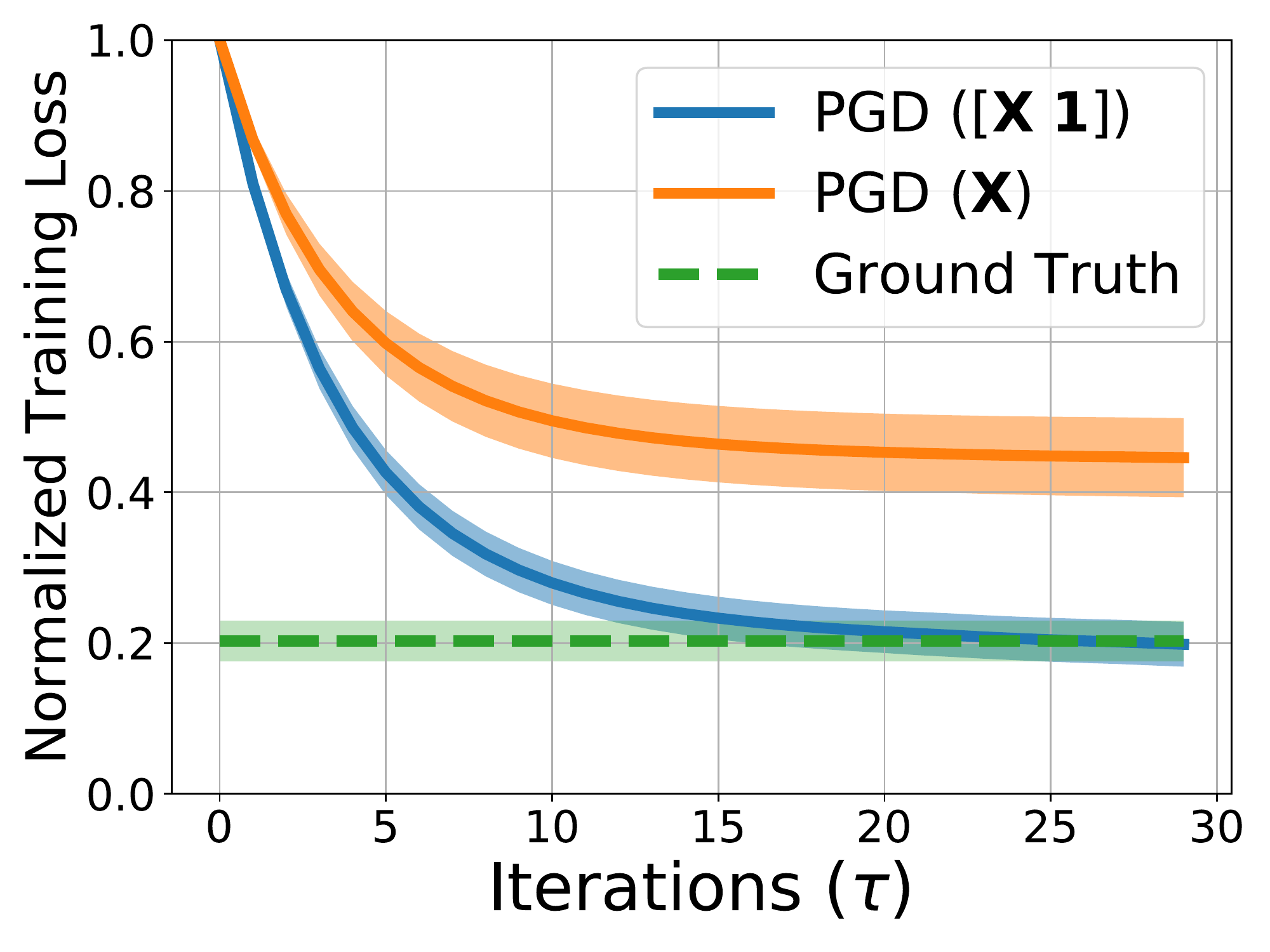}\vspace{-5pt}\subcaption{$\phi=\text{ReLU}$}\label{fig1c}
\end{subfigure}
\end{centering}~
\begin{centering}
\begin{subfigure}[t]{1.5in}
\includegraphics[height=0.8\linewidth,width=1\linewidth]{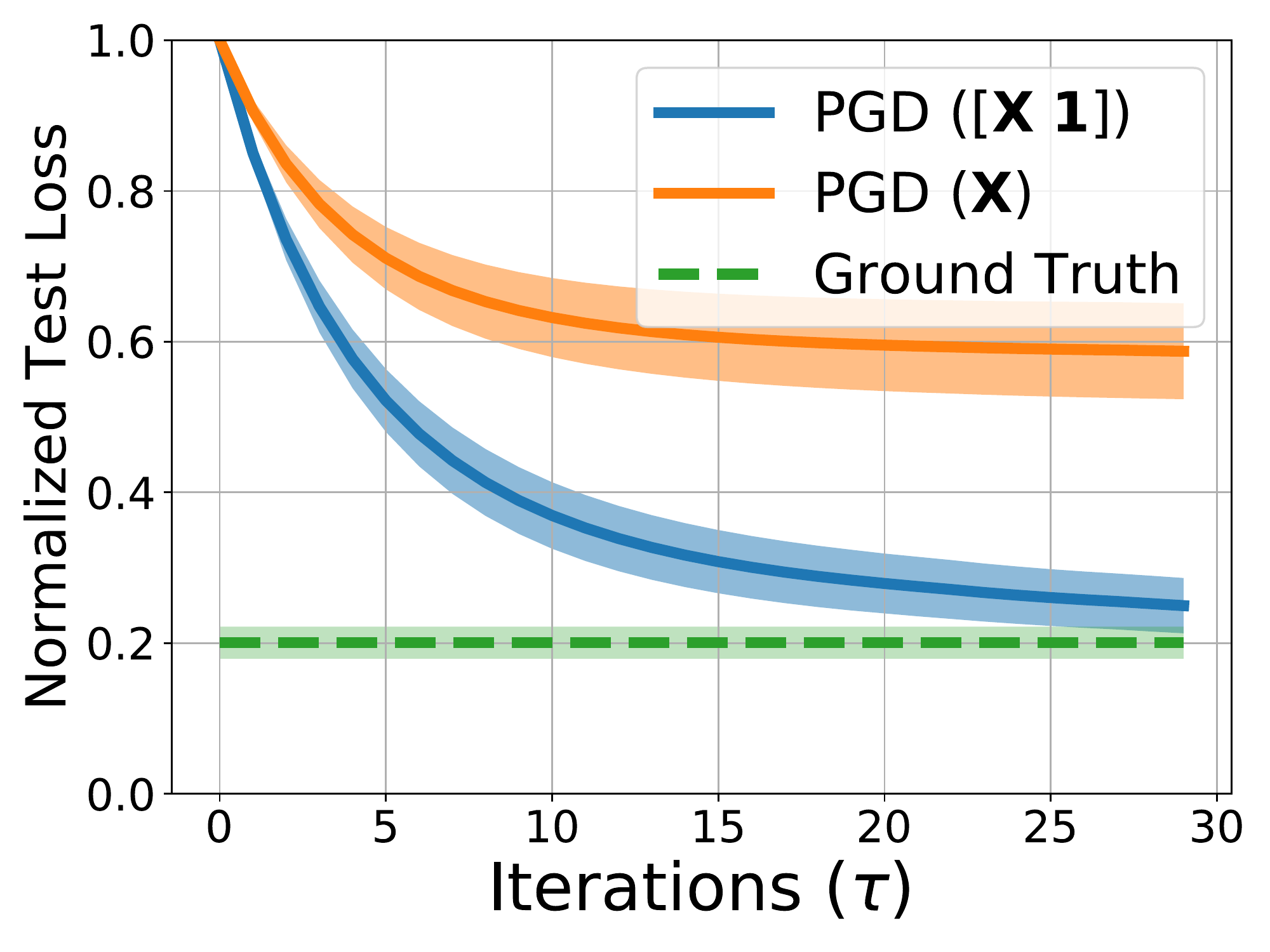}\vspace{-5pt}\subcaption{$\phi=\text{ReLU}$}\label{fig1d}
\end{subfigure}
\end{centering}
\vspace{-6pt}\caption{We run PGD with different activations (ReLU and sign) using $\X$ and $[\X~\co]$ as data matrices. In both cases, train and test errors decay gracefully to the ground truth baseline~(with debiased output). However, PGD using $[\X~\co]$ outperforms PGD using $\X$ alone for ReLU.}\label{fig1}
\end{figure}






In this section, we discuss experiments that corroborate our theoretical results. We consider a standard single-index model where for some ground truth vector $\bbeta$ and link function $\phi$, the input/output relation is given by $y_i=\phi(\bbeta^T\x_i)$. We pick $\bbeta$ to be a sparse vector with $s=20$ nonzeros and $p=800$ and set sample size to be $n=500$. Because of sparsity prior, we run PGD as iterative hard thresholding where $\bteta_{\tau}$ is projected to be $s$-sparse after every iteration. As link functions, we considered ReLU (i.e.~$\max(x,0)$) and sign functions (maps to $\pm 1$); which are of interest for deep learning and quantization respectively. We generate $\x_i$'s with i.i.d.~exponentially distributed entries (with parameter $\la=1$) and then remove the mean and normalize the covariance to identity. We pick a learning rate of $\eta=1/5n$ in all experiments. The shaded areas in the plots correspond to one standard deviation.

To assess test and training performance of PGD, we use the following three metrics:
\begin{itemize}
\item {\bf{the normalized training error}} defined as $\tn{\y-\X\bteta_{\tau}-\mu_{\tau}\co}^2/\tn{\y}^2$,
\item {\bf{the normalized test error}} that is similarly defined but evaluated on a fresh dataset of size $n$ using the training model $\bteta_{\tau}$,
\item {\bf{correlation to ground truth}} vector $\bbeta$ defined as $\frac{\bteta_{\tau}^T\bbeta}{\tn{\bteta_{\tau}}\tn{\bbeta}}$.
\end{itemize}
We compare two baselines. First one is running PGD with $\X$ and $[\X~\co]$ separately. Second one assumes knowledge of ground truth $\bbeta$ and fits a model $\gamma\bbeta$ by finding $\gamma$ to minimize the training loss. Numerically, we minimize $\tn{\bar{\y}-\gamma \X\bm{\beta}}^2$ over $\gamma$ where $\bar{y}_i=y_i-(1/n)\sum_{i=1}^n y_i$. This sets $\gamma=\bar{\y}^T\X\bm{\beta}/\tn{\bm{\X\beta}}^2$.

 Figure \ref{fig1} plots the loss as a function of the PGD iterations $\tau$. Both training and test errors gracefully decays with more iterations for both choices of link functions. The dashed values corresponds to $\gamma\bbeta$'s performance. While there is a slight mismatch between train/test performances (due to finite samples), high-dimensional estimation via PGD works well and performs on par with ground truth. Observe that for ReLU, $\E[y]$ is nonzero and estimating mean should be beneficial. Indeed, Figures \ref{fig1c} and \ref{fig1d} demonstrates that $[\X~\co]$ substantially outperforms using $\X$ alone. There is no improvement for sign function since $\E[y]\approx 0$.

\begin{figure}[t!]
\begin{centering}
\begin{subfigure}[t]{1.5in}
\includegraphics[height=0.8\linewidth,width=1\linewidth]{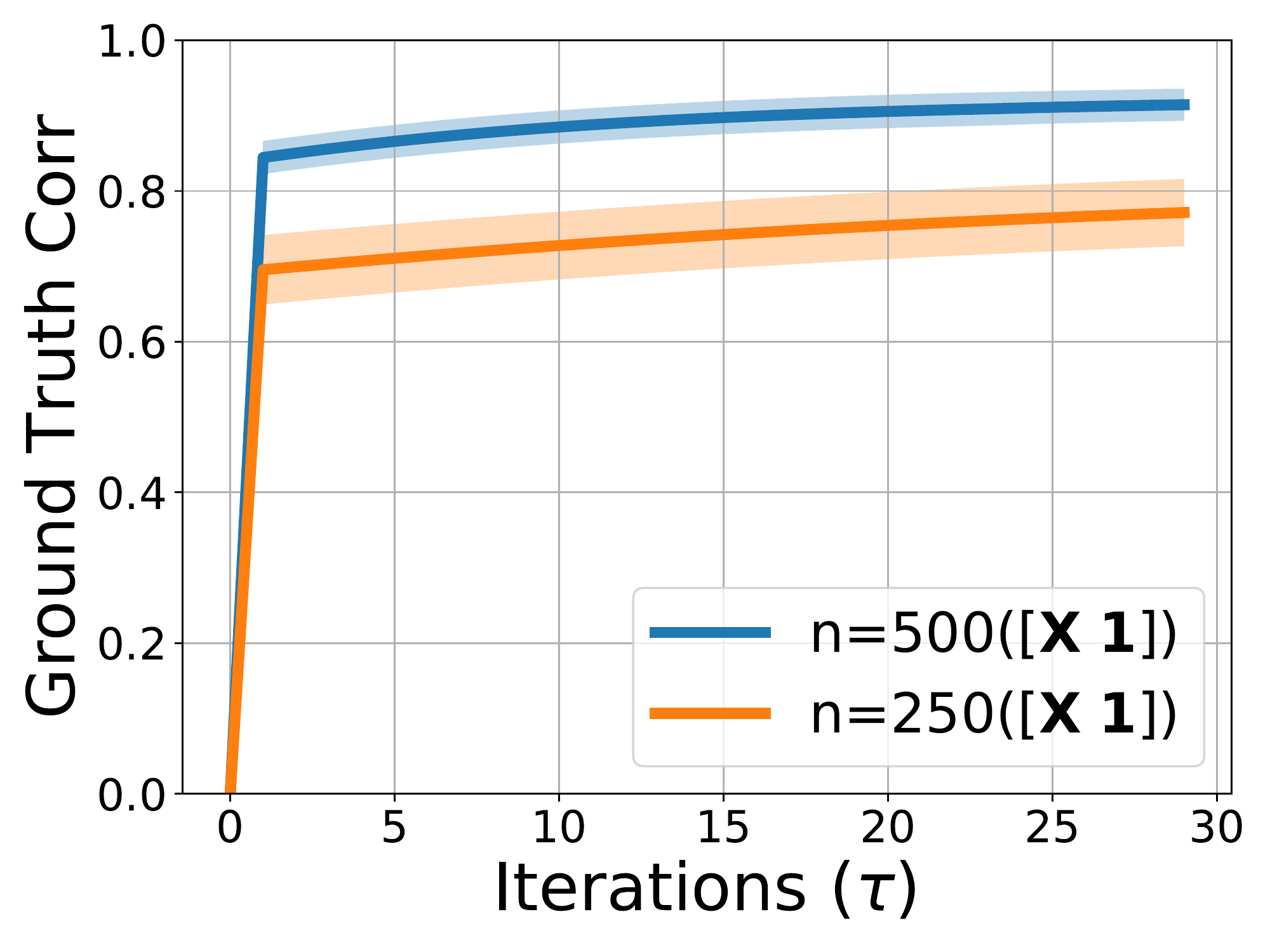}\vspace{-5pt}\subcaption{$\phi=\text{sign}$}\label{fig2a}
\end{subfigure}
\end{centering}~
\begin{centering}
\begin{subfigure}[t]{1.5in}
\includegraphics[height=0.8\linewidth,width=1\linewidth]{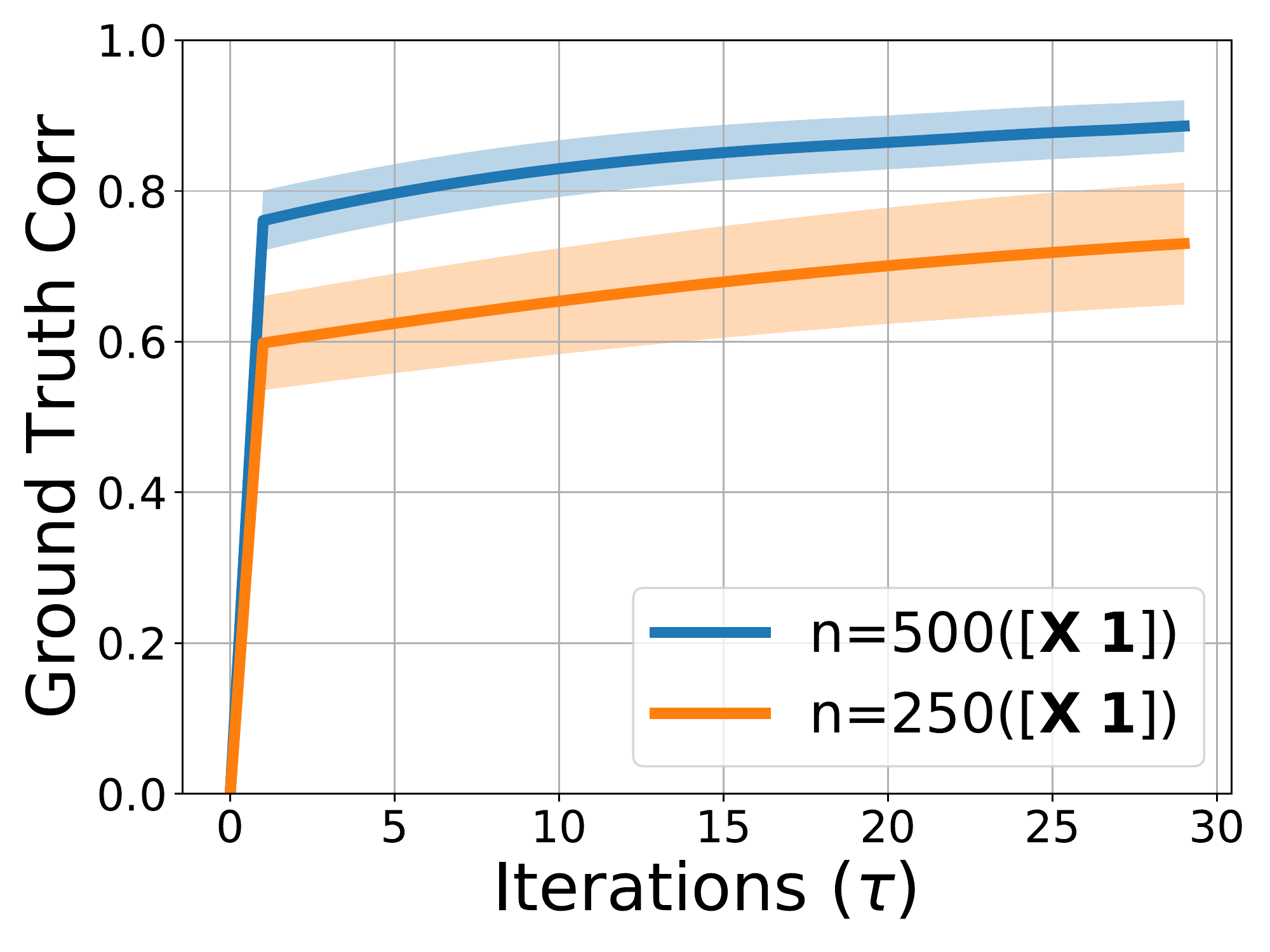}\vspace{-5pt}\subcaption{$\phi=\text{ReLU}$}\label{fig2b}
\end{subfigure}
\end{centering}~
\begin{centering}
	\begin{subfigure}[t]{1.5in}
		\includegraphics[height=0.8\linewidth,width=1\linewidth]{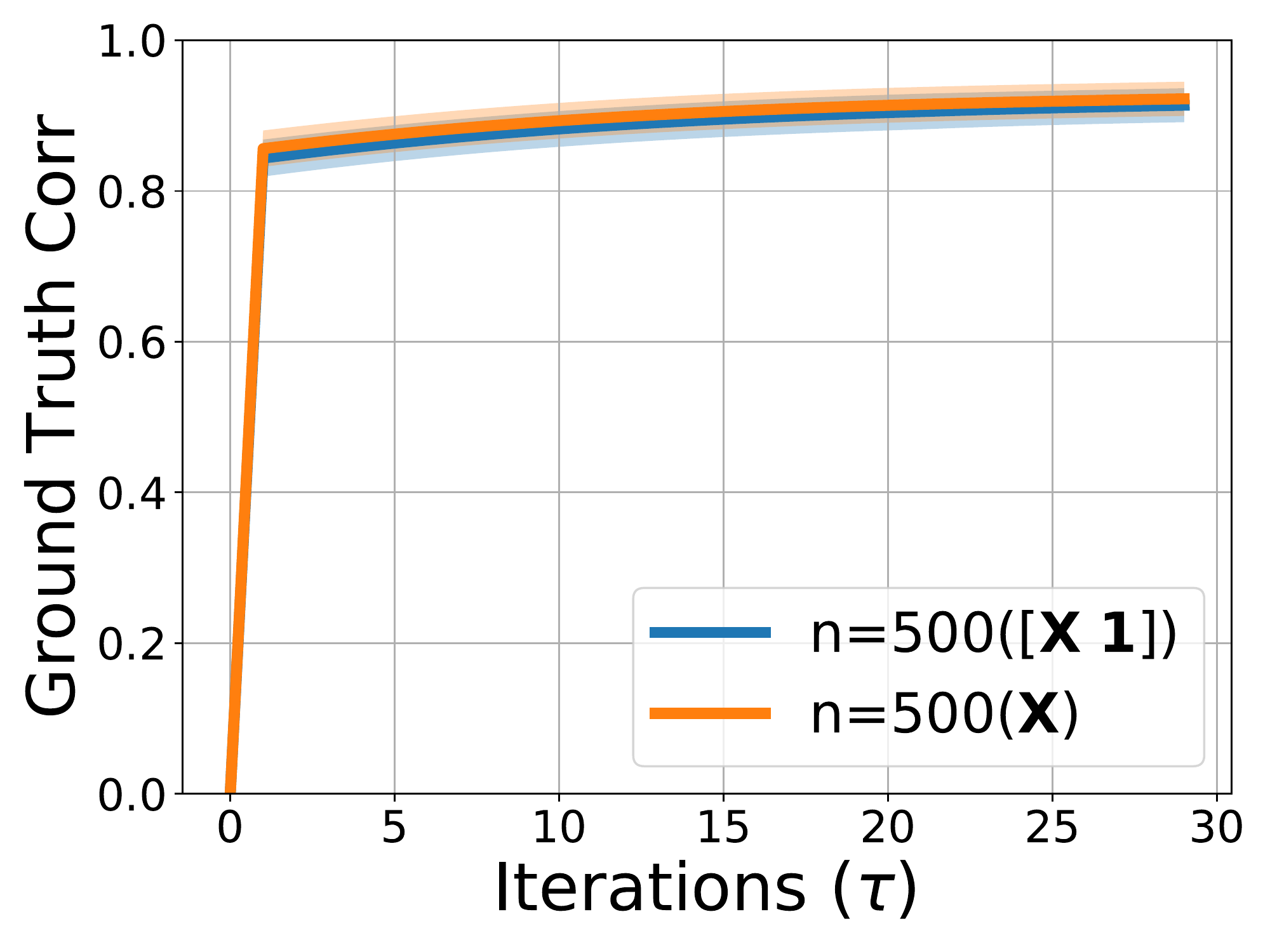}\vspace{-5pt}\subcaption{$\phi=\text{sign}$}\label{fig2c}
	\end{subfigure}
\end{centering}~
\begin{centering}
	\begin{subfigure}[t]{1.5in}
		\includegraphics[height=0.8\linewidth,width=1\linewidth]{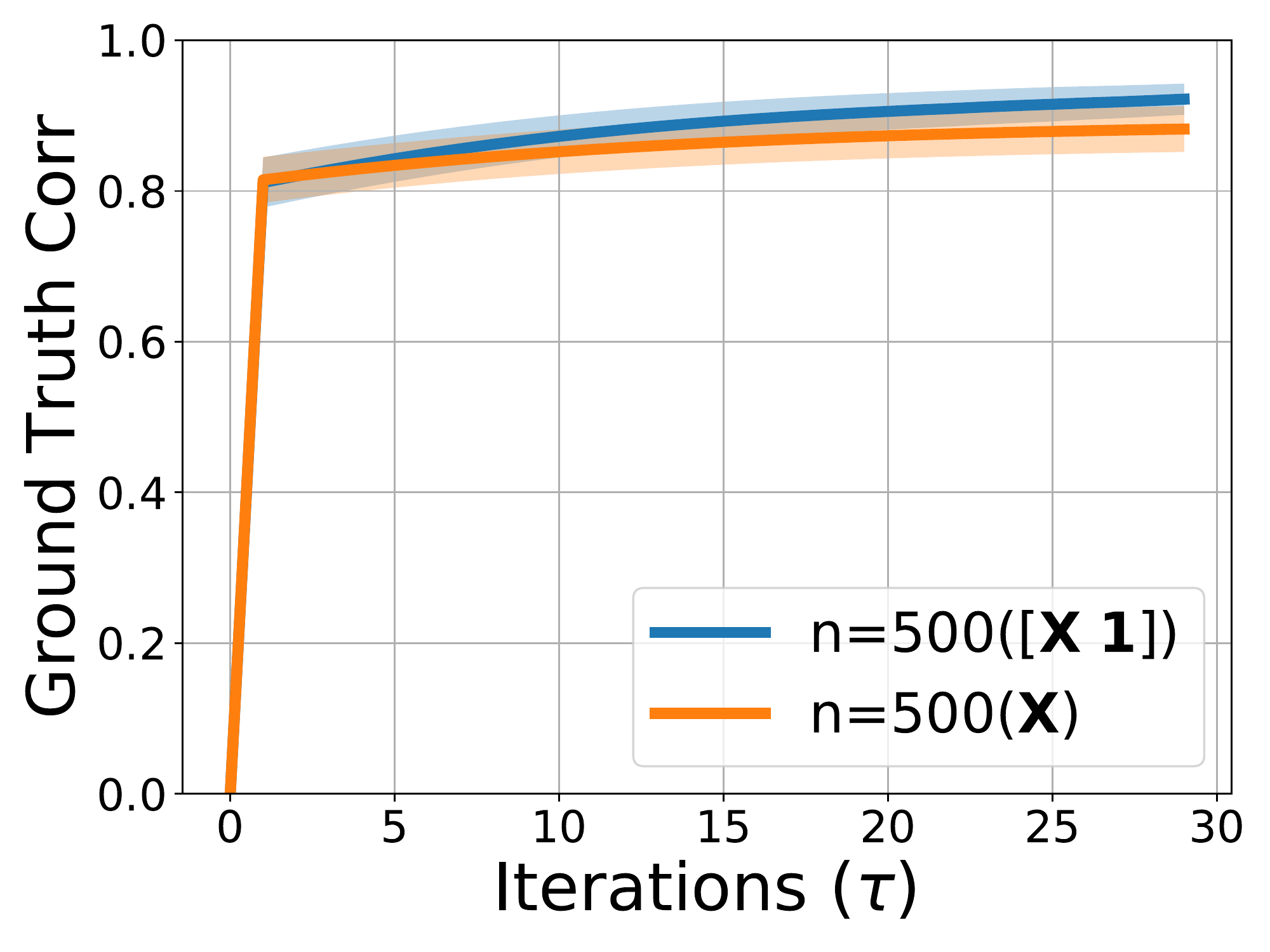}\vspace{-5pt}\subcaption{$\phi=\text{ReLU}$}\label{fig2d}
	\end{subfigure}
\end{centering}\vspace{-6pt}\caption{We plot correlation of PGD estimate with the ground truth vector $\bbeta$. Correlation increases with more samples and when we use $[\X~\co]$ instead of $\X$ in the PGD estimate.}\label{fig2}\vspace{-12pt}
\end{figure}

In Figure \ref{fig2} we focus on the parameter estimation question by plotting the correlation between $\bteta_{\tau}$ and $\bbeta$. Correlation is always between $-1,1$ and quantifies how well we can estimate direction of the ground truth vector via PGD. This experiment is conducted with two values of $n$ namely $250$ and $500$ while $p=800$ in both cases. Observe that, a larger sample size results in more stable estimation (smaller standard deviations) and higher correlation with output. Additionally Figure \ref{fig2d} shows that ReLU problem achieves better correlation once we account for the bias term. Hence, mean estimation is not only beneficial for test performance but also for parameter estimation.


\section{Proofs of Main Theorems}\label{sec: proof}
This section proves our main results and outlines the proofs of Theorems \ref{thrm:subgaussian_convergence}, \ref{thrm:subexp_convergence}, \ref{thrm:convergence_rates} and \ref{thrm:error_bound}. Throughout, we use the same notation as described in \ref{sec:math_prelim}.

\subsection{Proof of Theorem \ref{thrm:subexp_convergence}}
We provide our analysis for subexponential samples. The extension to subgaussian samples is accomplished in an identical fashion. Set the estimation error at iteration $\tau$ to be $\h_\tau=[{\bteta_{\tau}}^T~\mu_{\tau}]^T-[{\bteta^\star}^T~\mu^\star]^T$. Note that, when $\rho(\Cc) < 1$ and $\Rc$ is a convex regularizer, then the recursion \eqref{rec:main_recursion} can be iteratively expanded as
\begin{align}
\tn{\h_{\tau}} &\leq    \tn{\h_{0}}\rho(\Cc)^{\tau} +\eta \nu(\Cc)\sum_{k=0}^{\tau-1}\rho(\Cc)^k \nn \\
&\leq   \tn{\h_{0}}\rho(\Cc)^{\tau} +\eta \nu(\Cc)\sum_{k=0}^{\infty}\rho(\Cc)^k \nn \\
&=\tn{\h_{0}}\rho(\Cc)^{\tau} +\frac{\eta \nu(\Cc)}{1-\rho(\Cc)}\label{eqn: main_thrm_prf_eq1}
\end{align}
With the advertised probability, subexponential statements of Theorems \ref{thrm:convergence_rates} and \ref{thrm:error_bound} hold. Hence, for some constants, we have that $\rho(\Cc) \leq  1-c_0{\eta n}$, $\nu(\Cc) \leq C\sqrt{n}{\sigma(\omega_n(\Cc)+t)\log(n)}$ and $\eta={c}/{q}$ with $q=(n+p)\log^3(n+p)$. Plugging these in \eqref{eqn: main_thrm_prf_eq1}, we find the following upper bound on the right hand side,
\begin{align}
\tn{\h_{\tau+1}} &\leq (1-c_0{\eta n})^{\tau}\tn{\h_{0}} \nn \\
&~~+ \frac{\eta }{c_0{\eta n}} C{\sqrt{n}{\sigma(\omega_n(\Cc)+t)\log(n)}} \nn \\
&=  (1-\frac{c_0cn}{q})^{\tau}\tn{\h_{0}} \nn \\
&~~+ \sigma \frac{C}{c_0}\frac{(\omega_n(\Cc)+t)\log(n)}{\sqrt{n}}, \nn
\end{align}
which is the desired bound. The case of subgaussian samples is again a corollary of Theorems \ref{thrm:convergence_rates} and \ref{thrm:error_bound}. This concludes the proof of our main result.

\subsection{Proof of Theorem \ref{thrm:convergence_rates} for subgaussian samples}\label{sec: subgauss convergence}
We start our proof with the following lemma.
\begin{lemma}\label{lemma:general set behaves good}
Let $(\x_i)_{i=1}^n \sim \x \in \R^p$ be i.i.d.~isotropic subgaussian samples. Let $\X \in \R^{n\times p}$ be concatenated data and $[\X~\co]$ is the modified-data matrix, where $\co$ is a vector of all ones. Let $\Tc$ be a closed set with Euclidian radius bounded by a constant and 
\[
\Tc_{\text{ext}}=\{\vt\bgl \vt=[\beta \vb^T~\gamma]^T\}.
\] 
where $\beta\leq C_1,~\gamma\leq C_2$ for some positive constants $C_1,C_2$ and $\vb \in \Tc$. Assume $n \gtrsim (\omega(\Tc)+t)^2$. Then, with probability at least $1-2e^{-t^2}$ we have
\[
\sup_{\vt \in \Tc_{\text{ext}}} |\vt^T(\Iden-\frac{1}{n}[\X~\co]^T[\X~\co])\vt| \lesssim \frac{\omega(\Tc)+t}{\sqrt{n}}.
\]
\end{lemma}
The proof of Lemma~\ref{lemma:general set behaves good} is deferred to Section~\ref{subsec:proof of lemma general set behaves good}. Next using the result of Lemma~\ref{lemma:general set behaves good}, we obtain the following lemma which bounds the convergence rate for subgaussian samples.



\begin{lemma}\label{lemma: from uu to uv}
	Consider the setup of Lemma~\ref{lemma:general set behaves good}. Furthermore, let the tangent balls $\Cc$ and $\Cext$ be as defined in \eqref{def: tangent cone} and \eqref{def: ext tangent cone} respectively. Following Lemma~\ref{lemma:general set behaves good}, with probability at least $1-4e^{-t^2}$, the following holds 
	\[
	\sup_{\ut,\vt \in \Cext} |\ut^T(\Iden-\frac{1}{n}[\X~\co]^T[\X~\co])\vt| \lesssim \frac{\omega(\Cc)+t}{\sqrt{n}}.
	\] 
\end{lemma}
The proof of Lemma~\ref{lemma: from uu to uv} is deferred to Section~\ref{subsec: proof lemma uu to uv}. 
This completes the proof for subgaussian samples.
\subsection{Proof of Theorem \ref{thrm:convergence_rates} for subexponential samples}\label{sec: subexpo convergence}
Let $(\x_i)_{i=1}^n\sim\x \in \R^p$ be i.i.d.~isotropic subexponential vectors and $\X$ be the associated design matrix as previously. Let $\Cc$ and $\Cext$ be as defined in~\ref{def: tangent cone} and \ref{def: ext tangent cone} respectively. Assume $n \gtrsim \omega_n^2(\Cc)$. Our proof strategy is based on the observation that, we can bound the~(restricted) singular values of $[\X~\co]^T[\X~\co]$ with high probability for subexponential data as follows.
\subsubsection{Upper bounding the singular values}
In this section we will upper bound the largest eigenvalue of the matrix $[\X~\co]^T[\X~\co]$ with high probability. Towards this goal, we utilize Matrix Chernoff bound from \cite{tropp2015introduction}.
\begin{theorem}[Matrix Chernoff~\cite{tropp2015introduction}]\label{thrm:matrix_chernoff}
	Consider a finite sequence $\{\mtx{X}_i\}_{i=1}^n$ of independent, random, Hermitian matrices with common dimension $d$. Assume that
	\begin{align*}
	0 \leq \sigma_{\min}(\mtx{X}_i) \;\; \textrm{and} \;\; \|\mtx{X}_i\|\leq L \quad \textrm{for} \;\; i=1,\ldots,n
	\end{align*}
	Define the sum $\mtx{M}=\sum_{i=1}^n\mtx{X}_i$ and let $\zeta_{\max}$ be an upper bound on the spectral norm of the expectation $\E[\mtx{M}]$ i.e.~$
	\zeta_{\max} \geq \|\E[\mtx{M}]\| = \|\sum_{i=1}^n \E[\mtx{X}_i]\|
	$. 
	We have that
	\[
	\P\left( \|\mtx{M}\| \geq (1+\epsilon)\zeta_{\max}\right) \leq d\left[\frac{e^{\epsilon}}{(1+\epsilon)^{1+\epsilon}}\right]^{\frac{\zeta_{\max}}{L}} ,  \epsilon \geq 0.
	\]
\end{theorem}
We will use Theorem~\ref{thrm:matrix_chernoff} to bound the largest eigenvalue of $[\X~\co]^T[\X~\co]$. Observe that
\[
[\X~\co]^T[\X~\co]=\sum\limits_{i=1}^n \begin{bmatrix}\x_i\\1 \end{bmatrix}[\x_i^T~1].
\]
Clearly this matrix is positive semidefinite. To bound $\|[\x_i^T~1]^T[\x_i^T~1]\|$, we use the following lemma. 
\begin{lemma}[Spectral norm bound]~\label{lemma: spectral norm bound}
	Let $(\x_i)_{i=1}^n$ be i.i.d.~isotropic subexponential samples in $\R^p$. Then, with probability at least $1-2(n+p)^{-100}$ the spectral norm of all $\x_i\x_i^T$ matrices can be bounded as
	\[
	\|\x_i\x_i^T\| \leq \tn{\x_i}^2 \leq c p\log^2(n+p).
	\] 	
\end{lemma}
The proof of lemma~\ref{lemma: spectral norm bound} is deferred to Section~\ref{subsec: proof lemma spectral bound}. Lemma~\ref{lemma: spectral norm bound} guarantees that $\|[\x_i^T~1]^T[\x_i^T~1]\|\leq \tn{[\x_i^T~1]^T}^2=\tn{\x_i}^2+1 \leq Cp\log^2(n+p)$. Hence, we do satisfy the conditions required by Theorem~\ref{thrm:matrix_chernoff}. Before using Theorem~\ref{thrm:matrix_chernoff} we will upper bound the spectral norm of the expectation $\E[[\X~\co]^T[\X~\co]]$ as follows.
\begin{lemma}[Spectral norm bound of expectation]~\label{lemma: spectral norm bound expec}
	Let $\x \in \R^p$ be an isotropic subexponential vector, $\xt=[\x^T~1]^T$ and let $B=Cp\log^2(n+p)$ for sufficiently large constant $C>0$. Then we have
	\[
	\E\big[\xt\xt^T \bgl \tn{\xt}^2 \leq B\big] \preceq 2\Iden_p.
	\] 	
\end{lemma}  
The proof of Lemma~\ref{lemma: spectral norm bound expec} is deferred to Section~\ref{subsec: proof lemma spectral bound expec}. Thus, applying Lemma~\ref{lemma: spectral norm bound expec} on the set of all $[\x_i^T~1]^T$ satisfying $\|[\x_i^T~1]^T[\x_i^T~1]\| \leq C p\log^2(n+p)$, we find that with probability $1-2(n+p)^{-100}$ the following holds
\begin{align}
\|\E[[\X~\co]^T[\X~\co]]\|& =\|\E[\sum\limits_{i=1}^n \begin{bmatrix}\x_i\\1 \end{bmatrix}[\x_i^T~1]]\| \nn \\
& \leq \|\sum_{i=1}^n2\Iden_p\| \nn \\
&= 2n. \nn
\end{align}
Hence, we can pick $\zeta_{\max}\geq 2n$ to upper bound the largest eigenvalue of $\E[[\X~\co]^T[\X~\co]]$. Now, using Theorem~\ref{thrm:matrix_chernoff} with $\zeta_{\max} = C_0C(n+p)\log^3(n+p), L=Cp\log^2(n+p)$ and $\epsilon=e-1$ we get
\begin{align}
&\P\left( \|[\X~\co]^T[\X~\co]\| \geq eC_0C(n+p)\log^3(n+p)\right) \nn \\
&\quad\quad \quad \leq p\left[\frac{e^{e-1}}{e^e}\right]^{C_0 \frac{n+p}{p}\log(n+p)} \nn \\
&\quad\quad \quad =pe^{-C_0 \frac{n+p}{p}\log(n+p)}\leq (n+p)^{-100}.
\end{align}
Union bounding, with probability at least $1-3(n+p)^{-100}$,
\begin{align}
\|[\X~\co]^T[\X~\co]\| \lesssim  (n+p)\log^3(n+p).\label{eqn: RSV upper bound}
\end{align}
\subsubsection{Lower bounding the singular values}
In this section we will lower bound the gain of $[\X~\co]$ restricted to the tangent ball $\Cext$. We will utilize the notion of restricted singular value (RSV) to proceed.
\begin{definition}[Restricted singular value] \label{def:RSV} Given a matrix $\mtx{M}$ and a closed set $\Cc $, the RSV of $\M$ at $\Cc$ is defined as
	\[
	\sigma(\mtx{M},\Cc) = \min_{\bm{v}\in \Cc}\frac{\tn{\mtx{M}\bm{v}}}{\tn{\bm{v}}}.
	\]
\end{definition}
In the following, we will lower bound $\min_{\vt \in \Cext}\tn{[\X~\co]\vt}$ which is the RSV of $[\X~\co]$ at $\Cext$. 
Recall that any $\vt\in \Cext$ with unit Euclidian norm obeys $\vt=[\sqrt{1-\gamma^2}\vb^T~\gamma]^T \in \Cext$ for $|\gamma| \leq 1$ and $\vb \in \Cc$. Consequently
\begin{align}
&\tn{[\X~\co]\vt}^2 = \tn{\sqrt{1-\gamma^2}\X\vb+\gamma\co }^2 \nn \\
&~~~= (1-\gamma^2)~\tn{\X\vb}^2+\gamma^2\co^T\co +2\gamma \sqrt{1-\gamma^2} \co^T\X\vb  \nn \\
&~~~\geq (1-\gamma^2)~\tn{\X\vb}^2+\gamma^2n +2 \gamma\sqrt{1-\gamma^2}\vb^T\sum_{i=1}^n\x_i. \nn
\end{align}
Setting $\bar{\x}=\frac{1}{n}\sum_{i=1}^n \x_i$ and minimizing both sides over $\vt \in \Cext$, we get
\begin{align}
&\min_{\vt \in \Cext}\tn{[\X~\co]\vt}^2 \nn \\
&\quad \geq \min_{|\gamma|\leq 1}\big((1-\gamma^2)\min_{\vb \in \Cc}\tn{\X\vb}^2+\gamma^2n\big)-2 n\sup_{\vb \in \Cc}|\vb^T\bar{\x}|\nn \\
&  \quad\geq \min\{\min_{\vb \in \Cc}\tn{\X\vb}^2,n\}-2 n\sup_{\vb \in \Cc}|\vb^T\bar{\x}|\label{eqn inf_Xt to inf_X}.
\end{align}
In essence, \eqref{eqn inf_Xt to inf_X} bounds RSV of $[\X~\co]$ in terms of the RSV of $\X$ and some simpler terms. The following theorem from~\cite{oymak2018learning} (Theorem D.11) gives a lower lower bound on the RSV of a matrix $\X$ with i.i.d.~subexponential rows. 
\begin{theorem}[Bounding RSV~\cite{oymak2018learning}]\label{thrm:D.11}
	Let $\X \in \R^{n\times d}$ be a random matrix with i.i.d.~isotropic subexponential rows. Let $\Cc$ be a tangent ball as in \eqref{def: tangent cone} and suppose the sample size obeys $n \gtrsim (\omega_n(\Cc)+t)$. Then with probability at least $1-3\exp(-c \min(n,t\sqrt{n},t^2))$, we have that
	\[
	\min_{\vb \in \Cc} \tn{\X\vb}^2 \geq c_0 n.
	\]
\end{theorem}
Next, we shall state a lemma from~\cite{oymak2018learning} (Lemma D.7) to upper bound the term involving the sample average $\bar{\x}$.
\begin{lemma}[Bounding empirical width~\cite{oymak2018learning}]\label{lemma:D.7}
	Suppose $\Cc$ is a subset of the unit Euclidian ball and $(\x_i)_{i=1}^n$ are i.i.d.~zero-mean vectors with bounded subexponential norm. Define the empirical average vector $\bar{\x}=\frac{1}{n}\sum_i \x_i$. We have that
	\begin{align}
	&\P\left(\sup_{\ub \in \Cc}|\ub^T\bar{\x}|\leq C\frac{(\omega_n(\Cc)+t)}{\sqrt{n}}\right)\nn  \\
	&\quad \quad \quad \quad \quad \quad \geq 1-2\exp(-c~\min(t\sqrt{n}, t^2)). \nn
	\end{align}
\end{lemma}
Combining Theorem~\ref{thrm:D.11} and Lemma~\ref{lemma:D.7} into \eqref{eqn inf_Xt to inf_X} we find that, there exist constants $c,c_0,C_0>0$ such that with probability at least $1-5\exp(-c\min(n,t\sqrt{n},t^2))$, we can lower bound the RSV of $[\X~\co]$ as,
\begin{align}
\min_{\vt \in \Cext}\tn{[\X~\co]\vt}^2 &\geq c_0 n -C_0n\frac{\omega_n(\Cc)+t}{\sqrt{n}} \nn \\ 
&\geq c_0n/2\label{eqn: RSV lower bound}.
\end{align}
where last line follows from the assumption that $n\gtrsim (\omega_n(\Cc)+t)^2$.

\subsubsection{Upper bounding the convergence rate}
Union bounding the events \eqref{eqn: RSV upper bound} and \eqref{eqn: RSV lower bound}, we obtain upper and lower bounds on the singular values of $[\X~\co]$ with the desired probability. Hence, we can bound the convergence rate of PGD as follows. Setting $q=(n+p)\log^3(n+p)$, we have \eqref{eqn: RSV upper bound} $\|[\X~\co]^T[\X~\co]\| \leq C q$. Therefore, choosing learning rate $\eta={1/Cq}$, the matrix $\Iden-\eta [\X~\co]^T[\X~\co]$ is positive semidefinite~(PSD). Hence, applying the generalized Cauchy-Schwarz inequality for PSD matrix, we find
\begin{align}
\rho(\Cc)&=\sup_{\ut,\vt \in \Cext} \ut^T(\Iden-\eta[\X~\co]^T[\X~\co])\vt \nn \\
&\leq \sup_{\ut,\vt \in \Cext} [(\ut^T(\Iden-\eta[\X~\co]^T[\X~\co])\ut)^{1/2}\nn \\
& \quad \quad \quad \quad(\vt^T(\Iden-\eta[\X~\co]^T[\X~\co])\vt)^{1/2}] \nn \\
&=\sup_{\vt \in \Cext} \vt^T(\Iden-\eta[\X~\co]^T[\X~\co])\vt\nn \\
&=1-\eta \min_{\vt \in \Cext} \tn{[\X~\co]\vt}^2 \nn \\
&\leq 1-c_0\eta n/2.\nn
\end{align}
Here the last inequality follows from \eqref{eqn: RSV lower bound}. This completes the proof for subexponential samples.
\subsection{Proof of Theorem~\ref{thrm:error_bound} for subgaussian samples}\label{sec: subgauss error}
Suppose the dataset $(\x_i,y_i)_{i=1}^n \sim (\x,y)$ is $\sigma$-subgaussian. Let $\X, [\X~\co], \Cc$ and $\Cext$ be as defined in Section \ref{sec:math_prelim}, recall $\w$ from \eqref{residual vector} and assume $n \gtrsim (\omega(\Cc)+t)^2$. Representing $\vt\in\Cext$ as $\vt=[\sqrt{1-\gamma^2}\vb^T~\gamma]^T$ for $\vb\in\Cc$ and $|\gamma|\leq 1$, we have
\begin{align}
\nu(\Cc) &= \sup_{\vt \in \Cext}|\vt^T [\X~\co]^T \w|\nn \\
&= \sup_{\vb \in \Cc,|\gamma|\leq 1}|\sqrt{1-\gamma^2}\vb^T\X^T\w+\gamma\co^T\w| \nn \\
& \leq \sup_{\vb \in \Cc, |\gamma|\leq 1} |\sqrt{1-\gamma^2}\vb^T\X^T\w| + \sup_{|\gamma|\leq 1}|\gamma \co^T\w| \nn \\
& \leq \sup_{\vb \in \Cc} |\vb^T\X^T\w| + |\co^T\w|. \label{noise term split}
\end{align}
In the following we will upper bound the terms $\sup_{\vb \in \Cc} |\vb^T\X^T\w|$ and $| \co^T\w|$ separately and will combine them to get an upper bound on the residual error.
\subsubsection{Upper bounding the first term in \eqref{noise term split}}
In order to upper bound the first term in \eqref{noise term split}, define the clipping function
\[
\text{clip}(a,B)=\begin{cases}a~\text{if}~|a|\leq B\\\text{sign}(a)B~\text{else}\end{cases}.
\]
The following lemma immediately follows from union bounding the large deviations of subgaussian and subexponential variables~$X$ and shows that $X=\text{clip}(X,B)$ with high probability.
\begin{lemma}\label{lemma: w_clip=w}
Let $(w_i)_{i=1}^n$ be i.i.d.~subgaussian random variables with $\tsub{w_i} \leq \sigma$. There exists a constant $C>0$ such that picking $B=C\sqrt{\log(n)}$, with probability $1-2n^{-100}$ for all $i$, we have
\[
w_i=\text{clip}(w_i,\sigma B).
\]
If instead $(w_i)_{i=1}^n$ are i.i.d.~subexponential with $\te{w_i} \leq \sigma$, then picking $B=C\log(n)$ leads to the same result. 
\end{lemma}
Using Lemma \ref{lemma: w_clip=w}, $\|\w\|_{\infty}\leq \sigma B$ with probability $1-2n^{-100}$. Conditioned on this event, we have
\begin{align}
\sup_{\vb \in \Cc}|\vb^T\X^T\w| &= \sup_{\bm{v}\in \Cc}  |\bm{v}^T\sum_{i=1}^n \text{clip}(w_i,\sigma B)\bm{x_i}| \label{eqn res err subgauss pre}.
\end{align}
Setting $\z_i=\text{clip}(w_i,\sigma B)\bm{x_i}=w_i\x_i$, \eqref{eqn res err subgauss pre} can be re-written as
\begin{align}
\sup_{\vb \in \Cc}&|\vb^T\X^T\w| = \frac{1}{n}\sup_{\bm{v}\in \Cc}  |\bm{v}^T\sum_{i=1}^n \z_i|\nn\\
&\leq \sup_{\bm{v}\in \Cc}  |\bm{v}^T\sum_{i=1}^n (\z_i-\E[\z_i])|+ \sup_{\bm{v}\in \Cc}  |\bm{v}^T\sum_{i=1}^n\E[\z_i]|\nn\\
&\leq \sup_{\bm{v}\in \Cc}  |\bm{v}^T\sum_{i=1}^n (\z_i-\E[\z_i])|+n \tn{\E[\z_1]}.
\label{eqn res err subgauss}
\end{align}
Note that $\z_i=w_i\x_i$ is subgaussian since $w_i$ is bounded. The subgaussian norm obeys
\[
\tsub{\z_i-\E[\z_i]} \lesssim \tsub{\z_i} \lesssim \sigma \sqrt{\log(n)}\tsub{\x_i} \lesssim \sigma \sqrt{\log(n)}.
\]
Define the average vector $\bar{\z}=n^{-1/2}\sum_{i=1}^n(\z_i-\E[\z_i])$ which is still subgaussian with same norm (up to a constant). Standard results from functional analysis \cite{talagrand2014gaussian} guarantee 
\begin{align}
\frac{1}{n}\sup_{\bm{v}\in \Cc} | \bm{v}^T (\z_i&-\E[\z_i]) | = \frac{1}{\sqrt{n}} \sup_{\bm{v}\in \Cc} | \bm{v}^T \bar{\z} | \nn \\
&\lesssim \frac{\sigma(\omega(\Cc)+t)\sqrt{\log(n)}}{\sqrt{n}}. \label{eqn: empr width bound subgauss}
\end{align}
with probability at least $1-2e^{-t^2/2}$. This bounds the first term of \eqref{eqn res err subgauss}. Next, we address the expectation term $\tn{\E[\z_1]}$ via following lemma.
\begin{lemma}\label{thrm: bounding error exp}
	Suppose $\x$ is an isotropic Orlicz-a vector and $\|w\|_{\psi_a}\leq \sigma$. Let $B=C\log^{1/a}(n)$ for sufficiently large constant $C>0$. For $a=1,2$, we have that
	\[
	\tn{\E[w\x\bgl |w|\leq \sigma B]} \lesssim \sigma p^2n^{-201}.
	\]
\end{lemma}
The proof of Lemma~\ref{thrm: bounding error exp} is deferred to Section \ref{ref bound}. Combining \eqref{eqn: empr width bound subgauss} and Lemma~\ref{thrm: bounding error exp} into \eqref{eqn res err subgauss}, with probability at least $1-2e^{-t^2/2}-2n^{-100}$, we find that,
\begin{align}
\frac{1}{n}\sup_{\vb \in \Cc}|\vb^T\X^T\w| &\lesssim \frac{\sigma(\omega(\Cc)+t)\sqrt{\log(n)}}{\sqrt{n}}\nn \\
&~~+\sigma p^2n^{-200}\nn\\
&\lesssim \frac{\sigma(\omega(\Cc)+t)\sqrt{\log(n)}}{\sqrt{n}}
\label{eqn: uTXTw subgauss}
\end{align}
which is the desired bound for the first term in \eqref{noise term split}. 

\subsubsection{Upper bounding the second term in \eqref{noise term split}}
The vector $\w$ is zero-mean with $\tsub{\w}\leq \sigma$. Hence, $\tsub{\onebb^T\w}\leq \sigma\sqrt{n}$ which implies that with probability $1-2n^{-100}$, 
\[
|\onebb^T\w|\lesssim \sigma\sqrt{n\log n}.
\]
Combining the bound above with \eqref{eqn: uTXTw subgauss}, we get the advertised bound on the residual, namely
\begin{align}
\frac{1}{n}\nu(\Cc) \lesssim \frac{\sigma(\omega(\Cc)+t)\sqrt{\log(n)}}{\sqrt{n}},
\end{align}
with probability at least $1-2\exp(-t^2/2)-4n^{-100}$. This completes the proof for $\sigma$-subgaussian data.
\subsection{Proof of Theorem~\ref{thrm:error_bound} for subexponential samples}\label{sec: subexpo error}
Suppose the dataset $(\x_i,y_i)_{i=1}^n \sim (\x,y)$ is $\sigma$-subexponential. Let $\X, [\X~\co], \Cc$ and $\Cext$ be as defined in Section \ref{sec:math_prelim},  recall $\w$ from \eqref{residual vector} and assume $n \gtrsim (\omega_n(\Cc)+t)^2$. Similar to the subgaussian case, we split the residual into two terms via \eqref{noise term split} and bound each term separately to get a final bound.



\subsubsection{Upper bounding the first term in \eqref{noise term split}} Let $\z_i=w_i\x_i$. With probability $1-2n^{-100}$, we have that $\|\w\|_{\infty} \lesssim \sigma \log n$. We continue the analysis conditioned on this event. With bounded $w_i$, $\z_i-\E[\z_i]$ is subexponential via
\[
\te{\z_i-\E[\z_i]}\lesssim \te{\z_i}\lesssim \sigma \log n\te{\x_i}\lesssim \sigma\log n.
\]

Combining this with Lemma~\ref{lemma:D.7}, guarantees that 
\begin{align}
\frac{1}{n}\sup_{\vb \in \Cc}|\vb^T\sum_{i=1}^n(\z_i-\E[\z_i])| \lesssim \frac{\sigma(\omega_n(\Cc)+t)\log(n)}{\sqrt{n}} \label{eqn: empr width bound subexpo}
\end{align}
with probability at least $1-2\exp(-\order{\min(t\sqrt{n}, t^2)})$. 
Next, using Theorem~\ref{thrm: bounding error exp}, we also upper bound $\tn{\E[\z_1]}$ by $C\sigma p^2n^{-201}$. Combining this with \eqref{eqn: empr width bound subexpo} and substituting into (the deterministic inequality) \eqref{eqn res err subgauss}, with probability at least $1-2\exp(-\order{\min(t\sqrt{n}, t^2)})-2n^{-100}$ we have,
\begin{align}
\frac{1}{n}\sup_{\vb \in \Cc}|\vb^T\X^T\w|  \lesssim \frac{\sigma(\omega_n(\Cc)+t)\log(n)}{\sqrt{n}} .\label{eqn: uTXTw subexpo}
\end{align}

\subsubsection{Upper bounding the second term in \eqref{noise term split}}
Using $\te{w_i}\lesssim \sigma$ and applying Lemma \ref{lemma:D.7} (over one-dimensional $\R$), we find that $|\onebb^T\w|\lesssim \sigma(1+t)\sqrt{n}$ with probability $1-2\exp(-c\,\min(t\sqrt{n}, t^2))$.\\
\noindent Combining this with \eqref{eqn: uTXTw subexpo} and plugging into \eqref{noise term split}, we get the advertised upper bound 
\begin{align}
\frac{1}{n}\nu(\Cc) &\lesssim \frac{\sigma(\omega_n(\Cc)+t)\log(n)}{\sqrt{n}} + \frac{(1+t)\sigma}{\sqrt{n}} \nn \\
&\lesssim \frac{\sigma(\omega_n(\Cc)+t)\log(n)}{\sqrt{n}}
\end{align}
which holds with probability at least $1-4\exp(-c\,{\min(t\sqrt{n}, t^2)})-2n^{-100}$. This completes the proof for $\sigma$-subexponential data.


\section{Conclusion}\label{sec: conclusion}
We studied the problem of finding the best linear model from $n$ input-output samples under quadratic loss in the high-dimensional regime $n\ll p$.  For estimation, we utilized the projected gradient descent algorithm and showed its fast convergence as well as statistical accuracy in a data-dependent fashion. Our results are established for subexponential design which is heavier tailed compared to well-studied subgaussian. In both cases, we prove that {\em{nonlinearity of the problem behaves like independent noise}} and we establish favorable statistical guarantees as if the problem is linear. We also modified the original regression problem to allow for mean estimation and demonstrated its practical benefit when output labels have nonzero mean via simulations.

It would be desirable to extend our results to general loss function. If a loss function $\ell$ has the potential to better capture input/output relation, we can solve for\vspace{-5pt}
\[
\Lc(\bteta)=\sum_{i=1}^n\ell(y_i,\li\bteta,\x_i\ri).
\]
Specifically this function can still be quadratic but characterized by a nonlinear link function $\phi$ i.e.~$\ell(y_i,\li\bteta,\x_i\ri)=(y_i-\phi(\li\bteta,\x_i\ri))^2$. We believe that much of the results presented here extends to strongly-increasing $\phi$ where the derivative is lower bounded by a constant i.e.~$\phi'\geq \alpha$ for some $\alpha>0$. These functions are shown to behave like linear regression \cite{oymak2018stochastic}. However, it is not immediately clear if strong statistical and computational guarantees established in this paper (as well as related literature) can be established for $\phi$.

\section{Appendix}
This section provides the proofs of supporting results.
\subsection{Proof of Lemma~\ref{lemma:general set behaves good}} \label{subsec:proof of lemma general set behaves good}
We start by expanding the convergence term by substituting $\vt=[\beta\vb^T~\gamma]^T$ as follows,
\begin{align}
&|\vt^T(\Iden-\frac{1}{n}[\X~\co]^T[\X~\co])\vt| \nn \\
&= |\frac{1}{n}\tn{[\X~\co]\vt}^2-\tn{\vt}^2| \nn \\
&= |\frac{1}{n}\tn{\beta \X \vb +\gamma \co}^2-\tn{[\beta \vb^T~\gamma]^T}^2| \nn \\
&=|\frac{1}{n}(\beta^2\tn{\X\vb}^2+\gamma^2\co^T\co+2\beta \gamma\co^T\X\vb)\nn -\beta^2\tn{\vb}^2 -\gamma^2| \nn \\ 
&=|\frac{1}{n}\beta^2\tn{\X\vb}^2-\beta^2\tn{\vb}^2+\frac{1}{n}\gamma^2n-\gamma^2+2\beta \gamma\frac{1}{n}\sum_{i=1}^n\vb^T\x_i \nn \\
&\leq \beta^2|\frac{1}{n}\tn{\X\vb}^2-\tn{\vb}^2| +|2\beta \gamma||\vb^T\frac{\sum_{i=1}^n\x_i}{n}| \nn \\
&\lesssim |\vb^T(\Iden-\frac{1}{n}\X^T\X)\vb| +|\vb^T \bar{\x}|, \label{eqn conv Xt to X}
\end{align} 
where, $\bar{\x} = n^{-1}\sum_{i=1}^n\x_i$ is the empirical average vector of i.i.d. subgaussian rows $(\x_i)_{i=1}^n$. Thus, using \eqref{eqn conv Xt to X}, we can write 
\begin{align}
\sup_{\vt \in \Tc_{\text{ext}}} &|\vt^T(\Iden-\frac{1}{n}[\X~\co]^T[\X~\co])\vt| \nn \\
&\lesssim \sup_{\vb \in \Tc} |\vb^T(\Iden-\frac{1}{n}\X^T\X)\vb| +\sup_{\vb \in \Tc}|\vb^T \bar{\x}| . \label{eqn conv Xt two terms}
\end{align}
Given  $\X \in \R^{n\times p}$ is isotropic subgaussian, Lemma 6.14 in \cite{oymak2015sharp} guarantees
\begin{align} \label{eqn: ISG result}
\sup_{\vb \in \Tc} |\vb^T(\Iden-\frac{1}{n}\X^T\X)\vb| \lesssim {\frac{\omega(\Tc)+t}{\sqrt{n}}},
\end{align}
with probability at least $1-e^{-t^2}$. Furthermore, since $(\x_i)_{i=1}^n$'s have bounded subgaussian norm, $\bar{\x}$ is also bounded and standard results from functional analysis guarantee \cite{talagrand2014gaussian} 
\begin{align}\label{eqn: bounding emp width subgauss}
\sup_{\vb \in \Tc} | \vb^T \frac{\sum_{i=1}^n\x_i}{n} |= \sup_{\vb \in \Tc} | \vb^T \bar{\x} |  \lesssim{\frac{\omega(\Tc)+t}{\sqrt{n}}},
\end{align}
with probability at least $1-e^{-t^2}$. Combining the results \eqref{eqn: ISG result} and \eqref{eqn: bounding emp width subgauss} into \eqref{eqn conv Xt two terms}, we find that
\begin{align}\label{eqn: subgauss bound uu}
\sup_{\vt \in \Tc_{\text{ext}}} |\vt^T(\Iden-&\frac{1}{n}[\X~\co]^T[\X~\co])\vt| \lesssim \frac{\omega(\Tc)+t}{\sqrt{n}}
\end{align}
holds with probability at least $1-2e^{-t^2}$. This completes the proof of Lemma~\ref{lemma:general set behaves good}
\subsection{Proof of Lemma~\ref{lemma: from uu to uv}}\label{subsec: proof lemma uu to uv}
Let the tangent balls $\Cc$ and $\Cext$ be as defined in \eqref{def: tangent cone} and \eqref{def: ext tangent cone} respectively. Define the sets
\[
\Tc_-=\Cext-\Cext
\quad \textrm{and} \quad \Tc_+=\Cext+\Cext
\]
and note that,
\begin{align*}
\omega(\Cc-\Cc)&=\E[\sup_{\ub,\vb \in \Cc} \g^T(\ub-\vb)] \nn \\
&\leq \E[\sup_{\ub \in \Cc} \g^T\ub + \sup_{\vb \in -\Cc} \g^T \vb] = 2 \omega(\Cc).
\end{align*}
Similarly, $\omega(\Cc+\Cc) \leq 2 \omega(\Cc)$. Applying Lemma \ref{lemma:general set behaves good} on $\Tc_+$ and $\Tc_-$, with advertised probability, we have
\[
\sup_{\ab\in \Tc_+\cup \Tc_-} |\Lambda(\ab,\ab)|\lesssim \frac{\omega(\Cc)+t}{\sqrt{n}}.
\]
where $\Lambda(\ab,\bb)=\ab^T(\Iden-\frac{1}{n}[\X~\co]^T[\X~\co])\bb$. Now, for any $\ub,\vb\in \Cext$, picking $\ub+ \vb\in \Tc_+~\text{and}~\ub-\vb\in \Tc_-$, we have
\[
|\La(\ub+ \vb,\ub+ \vb)|,|\La(\ub-\vb,\ub-\vb)|\lesssim \frac{\omega(\Cc)+t}{\sqrt{n}}.
\]
To proceed, note that
\[
\La(\ub,\vb)= \frac{\La(\ub+ \vb,\ub+ \vb)-\La(\ub- \vb,\ub-\vb)}{4}.
\]
Hence, $|\La(\ub,\vb)|= |\ub^T(\Iden-\frac{1}{n}[\X~\co]^T[\X~\co])\vb| \lesssim(\omega(\Cc)+t)/{\sqrt{n}}$ holds with the advertised probability.

\subsection{Proof of Lemma~\ref{lemma: spectral norm bound}}\label{subsec: proof lemma spectral bound}
Let $(\x_i)_{i=1}^n\sim\x \in \R^p$ be i.i.d.~isotropic subgaussian samples and $\X \in \R^{n\times p}$ is the concatenated design matrix. Let $x_{ij}$ denotes the $ij^{th}$ element of the matrix $\X$. Since each $x_{ij}$ has subexponential norm bounded by a constant, there exists a constant $C>0$ such that  $|x_{ij}| \leq C\log(n+p)$ holds with probability at least $1-2(n+p)^{-102}$ using subexponential tail bound. Union bounding over all entries of $\X$ yields that $|x_{ij}| \leq C\log(n+p)$ holds for all $i,j$ with probability at least $1-2(n+p)^{-100}$. Hence, we can bound each row $\x_i$ of $\X$ with probability at least $1-2(n+p)^{-100}$ via
\begin{align} \label{row_norm_bound}
\tn{\x_i} \leq C\sqrt{p}\log(n+p),
\end{align}
or equivalently, we have 
\[
 \|\x_i\x_i^T\| \leq \tn{\x_i}^2 \leq cp\log^2(n+p).
\]
This completes the proof of Lemma~\ref{lemma: spectral norm bound}. 

\subsection{Proof of Lemma~\ref{lemma: spectral norm bound expec}}\label{subsec: proof lemma spectral bound expec}
Recall that $(\x_i)_{i=1}^n\sim \x \in \R^p$ are i.i.d.~isotropic subexponential vectors and $\xt=[\x^T~1]^T$. We can estimate the covariance matrix of $\xt$ given $\tn{\xt}^2\leq B$ using law of total probability as follows
\begin{align}
\E\big[\xt\xt^T\big] &=\E\big[\xt\xt^T \bgl \tn{\xt}^2\leq B\big]\P\big(\tn{\xt}^2\leq B\big)\nn \\
&+\E\big[\xt\xt^T \bgl \tn{\xt}^2> B\big]\P\big(\tn{\xt}^2> B\big) \label{eqn: covar expansion}.
\end{align}
Since a covariance matrix is positive-semidefinite, each term in \eqref{eqn: covar expansion} is individually positive semidefinite. Hence, we will drop the second term in \eqref{eqn: covar expansion} to get the following lower bound on the covariance matrix
\begin{align}
\hspace{-10pt}\E[\xt\xt^T] &\succeq \E[\xt\xt^T \bgl \tn{\xt}^2\leq B]\P(\tn{\xt}^2\leq B) \label{eqn: covar upper bound}
\end{align}
Using Lemma~\ref{lemma: spectral norm bound}, it follows that $\tn{\xt}^2=\tn{[\x^T~1]^T}^2\leq  Cp\log^2(n+p)=B$ holds with probability at least $1-2(n+p)^{-100}$. Hence, following \eqref{eqn: covar upper bound}, we get
\begin{align}
\E\big[\xt\xt^T \bgl \tn{\xt}^2 \leq B\big]& \preceq \frac{\E\big[\xt\xt^T\big]}{\P\big(\tn{\xt}\leq B\big)} \nn\\
&\preceq \frac{1}{1-2(n+p)^{-100}}\Iden_p \preceq 2\Iden_p.\nn
\end{align}
This completes the proof of Lemma~\ref{lemma: spectral norm bound expec}.

\subsection{Proof of Lemma~\ref{lemma: w_clip=w}}
\noindent{\bf{Subgaussian case:}} Using subgaussian tail, for large enough constant $C>0$, for each $i$, we have $|w_i| \leq C\sigma\sqrt{\log(n)}=\sigma B$ with probability at least $1-2n^{-101}$.  This implies $\text{clip}(w_i,\sigma B)=w_i$. Union bounding over all entries of $\w$, we find the result which holds with probability at least $1-2n^{-100}$.

\noindent {\bf{Subexponential case:}} follows similarly with $B=C \log(n)$. 

\subsection{Proof of Lemma~\ref{thrm: bounding error exp}}\label{ref bound}
We prove the result for subexponential samples. Subgaussian case follows similarly. Without loss of generality, let $\sigma=1$ as everything can be scaled accordingly. Defining clip function as previously, set $\z=\text{clip}(w, B)\x$. Furthermore, let $\wtail$ denotes the tail of $|w|$, such that,
\begin{align}\label{eqn: w_tail}
\wtail = \begin{cases} |w| \quad \text{if} \;\; |w|> B \\
0 \quad \text{otherwise}
\end{cases}.
\end{align} 
$\wtail$ is an upper bound on the error due to clipping, that is,
\begin{align}\label{eqn: w_tail as upper bound}
|w-\text{clip}(w, B)| \leq \wtail.
\end{align}

We proceed by upper bounding $\tn{\E[\z]}$ in terms of $\wtail$, using subadditive property of $\ell_2$-norm and the orthogonality of $w$ and $\x$~(i.e., $\E[w\x]=0$) as follows
\begin{align}
\tn{\E[\z]} &= \tn{\E[\text{clip}(w, B)\x]} \nn \\
&= \tn{\E[ (w-\text{clip}(w, B))\x]} \nn \\
&\leq \E[|w-\text{clip}(w, B)|\tn{\x}]  \nn \\
&\leq \E[\wtail\max(\tn{\x},\sqrt{p}B)]. \label{bound on average}
\end{align}
Using subexponentiality, for some constant $c>0$, we have that, $\P(\wtail > \sqrt{c} t) \leq 2e^{-t}$ and $\P(\tn{\x} > \sqrt{cp}t) \leq 2pe^{-t}$, where, the latter follows from union bounding over all entries of $\x$. Union bounding these two events, we get the following tail bound for their product,
\begin{align}\label{prod tail bound}
\P(\wtail\tn{\x}>c\sqrt{p}t^2) \leq 4pe^{-t}.
\end{align}
For notational convenience, set
\begin{align}\label{eqn:def_of_g}
g=\wtail\max(\tn{\x},\sqrt{p}B),
\end{align}
and note that $g$ satisfies the following property due to \eqref{eqn: w_tail}
\begin{align} \label{def of g}
&\begin{cases}
\text{either} \quad &g> \sqrt{p} B^2\\
\text{or} \quad &g=0 
\end{cases}.
\end{align}
Furthermore, from \eqref{prod tail bound} we get the following tail distribution
\begin{align} \label{eqn:tail dist g}
Q_g(t)=\P(g>t) \leq 4pe^{-[\frac{t}{c \sqrt{p}}]^{1/2}}.
\end{align}
for $t\geq \alpha:=\sqrt{p}B^2$. Combining~\eqref{eqn:def_of_g},~\eqref{def of g} and~\eqref{eqn:tail dist g}  into \eqref{bound on average} and denoting probability density function of $g$ by $f_g$, we get
\begin{align}
\tn{\E[\z]} &\leq \E[g] 
=\int_{\alpha}^{\infty}t f_g(t) dt 
=-\int_{\alpha}^{\infty}tdQ_g(t) \nn \\
& = -t Q_g(t)\big|_{\alpha}^{\infty}+\int_{\alpha}^{\infty}Q_g(t)dt \nn \\
& = \sqrt{p} B^2 Q_g( \sqrt{p} B^2) +\int_{\alpha}^{\infty}Q_g(t)dt \nn \\
& \leqsym{a} 4p^2 B^2e^{-B/\sqrt{c}} + 4p\int_{ \sqrt{p}B^2}^{\infty}e^{-[\frac{t}{ c \sqrt{p}}]^{1/2}} dt.\label{Q itegral}
\end{align}
where, (a) follows from~\eqref{eqn:tail dist g}. To bound the term on the right hand side, we do a change of variable in \eqref{Q itegral} by setting $\tau = [t/(c\sqrt{p})]^{1/2}$ to get,
\begin{align}
4p\int_{ \sqrt{p}B^2}^{\infty}e^{-[\frac{t}{ c \sqrt{p}}]^{1/2}} dt& \leq  8cp^2\int_{\frac{B}{\sqrt{c}}}^{\infty}\tau e^{-\tau} d \tau \nn \\
&\leq  8c p^2 \big[-\tau e^{-\tau}\big|_{\frac{B}{\sqrt{c}}}^{\infty} + \int_{\frac{B}{\sqrt{c}}}^{\infty}e^{-\tau} d \tau \big] \nn \\
&= 8 c p^2 \big[\frac{B}{\sqrt{c}} e^{-\frac{B}{\sqrt{c}}} + e^{-\frac{B}{\sqrt{c}}} \big] \nn \\
&\leq  8 c p^2 (\frac{B}{\sqrt{c}}+1)e^{-\frac{B}{\sqrt{c}}}.\label{final bound on subexpo ito B}
\end{align}
Combining this with~\eqref{Q itegral}, we get
\begin{align*}
\tn{\E[\z]} &\leq 4 p^2 (B^2+2 c(B/\sqrt{c}+1))e^{-B/\sqrt{c}} \\
& \leqsym{a} C_0 p^2n^{-201},
\end{align*}
where, we get (a) by picking $B=C\log(n)$ with sufficiently large $C>0$. Finally, note that conditioned on $|w|\leq B$, $\z=w\x$ and
\[
\tn{\E[\z]}\geq \tn{\E[w\x\bgl |w|\leq  B]}\Pro(|w|\leq B).
\]
Since $\Pro(|w|\leq B)>1/2$, this yields $ \tn{\E[w\x\bgl |w|\leq  B]}\lesssim{p^2n^{-201}}$ which is the advertised result with $\sigma=1$.

Similarly for subgaussian samples, one can show that 
\begin{align}
\tn{\E[\z]} \lesssim  p^2 B^2 e^{-B^2/c}.\label{final bound on subgauss ito B}
\end{align}
Picking $B=C\sqrt{\log(n)}$ with sufficiently large $C>0$, we get the same result, concluding the proof of Lemma~\ref{thrm: bounding error exp}.

\small{
	\bibliographystyle{IEEEtran}
	\bibliography{Bibfiles}

\begin{thebibliography}{10}
\providecommand{\url}[1]{#1}
\csname url@samestyle\endcsname
\providecommand{\newblock}{\relax}
\providecommand{\bibinfo}[2]{#2}
\providecommand{\BIBentrySTDinterwordspacing}{\spaceskip=0pt\relax}
\providecommand{\BIBentryALTinterwordstretchfactor}{4}
\providecommand{\BIBentryALTinterwordspacing}{\spaceskip=\fontdimen2\font plus
\BIBentryALTinterwordstretchfactor\fontdimen3\font minus
  \fontdimen4\font\relax}
\providecommand{\BIBforeignlanguage}[2]{{%
\expandafter\ifx\csname l@#1\endcsname\relax
\typeout{** WARNING: IEEEtran.bst: No hyphenation pattern has been}%
\typeout{** loaded for the language `#1'. Using the pattern for}%
\typeout{** the default language instead.}%
\else
\language=\csname l@#1\endcsname
\fi
#2}}
\providecommand{\BIBdecl}{\relax}
\BIBdecl

\bibitem{plan2016high}
Y.~Plan, R.~Vershynin, and E.~Yudovina, ``High-dimensional estimation with
  geometric constraints,'' \emph{Information and Inference: A Journal of the
  IMA}, vol.~6, no.~1, pp. 1--40, 2016.

\bibitem{boufounos20081}
P.~T. Boufounos and R.~G. Baraniuk, ``1-bit compressive sensing,'' in
  \emph{Information Sciences and Systems, 2008. CISS 2008. 42nd Annual
  Conference on}.\hskip 1em plus 0.5em minus 0.4em\relax IEEE, 2008, pp.
  16--21.

\bibitem{ganti2015learning}
R.~Ganti, N.~Rao, R.~M. Willett, and R.~Nowak, ``Learning single index models
  in high dimensions,'' \emph{arXiv preprint arXiv:1506.08910}, 2015.

\bibitem{oymak2017fast}
S.~Oymak and M.~Soltanolkotabi, ``Fast and reliable parameter estimation from
  nonlinear observations,'' \emph{SIAM Journal on Optimization}, vol.~27,
  no.~4, pp. 2276--2300, 2017.

\bibitem{plan2017high}
Y.~Plan, R.~Vershynin, and E.~Yudovina, ``High-dimensional estimation with
  geometric constraints,'' \emph{Information and Inference: A Journal of the
  IMA}, vol.~6, no.~1, pp. 1--40, 2017.

\bibitem{plan2016generalized}
Y.~Plan and R.~Vershynin, ``The generalized lasso with non-linear
  observations,'' \emph{IEEE Transactions on information theory}, vol.~62,
  no.~3, pp. 1528--1537, 2016.

\bibitem{thrampoulidis2015lasso}
C.~Thrampoulidis, E.~Abbasi, and B.~Hassibi, ``Lasso with non-linear
  measurements is equivalent to one with linear measurements,'' in
  \emph{Advances in Neural Information Processing Systems}, 2015, pp.
  3420--3428.

\bibitem{jacques2013robust}
L.~Jacques, J.~N. Laska, P.~T. Boufounos, and R.~G. Baraniuk, ``Robust 1-bit
  compressive sensing via binary stable embeddings of sparse vectors,''
  \emph{IEEE Transactions on Information Theory}, vol.~59, no.~4, pp.
  2082--2102, 2013.

\bibitem{vershynin2015estimation}
R.~Vershynin, ``Estimation in high dimensions: a geometric perspective,'' in
  \emph{Sampling theory, a renaissance}.\hskip 1em plus 0.5em minus 0.4em\relax
  Springer, 2015, pp. 3--66.

\bibitem{dirksen2017one}
S.~Dirksen, H.~C. Jung, and H.~Rauhut, ``One-bit compressed sensing with
  partial gaussian circulant matrices,'' \emph{arXiv preprint
  arXiv:1710.03287}, 2017.

\bibitem{dirksen2018robust_circ}
S.~Dirksen and S.~Mendelson, ``Robust one-bit compressed sensing with partial
  circulant matrices,'' \emph{arXiv preprint arXiv:1812.06719}, 2018.

\bibitem{plan2013robust}
Y.~Plan and R.~Vershynin, ``Robust 1-bit compressed sensing and sparse logistic
  regression: A convex programming approach,'' \emph{Information Theory, IEEE
  Transactions on}, vol.~59, no.~1, pp. 482--494, 2013.

\bibitem{agarwal2010fast}
A.~Agarwal, S.~Negahban, and M.~J. Wainwright, ``Fast global convergence rates
  of gradient methods for high-dimensional statistical recovery,'' in
  \emph{Advances in Neural Information Processing Systems}, 2010, pp. 37--45.

\bibitem{oymak2015sharp}
S.~Oymak, B.~Recht, and M.~Soltanolkotabi, ``Sharp time--data tradeoffs for
  linear inverse problems,'' \emph{IEEE Transactions on Information Theory},
  vol.~64, no.~6, pp. 4129--4158, 2018.

\bibitem{giryes2018tradeoffs}
R.~Giryes, Y.~C. Eldar, A.~M. Bronstein, and G.~Sapiro, ``Tradeoffs between
  convergence speed and reconstruction accuracy in inverse problems,''
  \emph{IEEE Transactions on Signal Processing}, vol.~66, no.~7, pp.
  1676--1690, 2018.

\bibitem{beck2009fast}
A.~Beck and M.~Teboulle, ``A fast iterative shrinkage-thresholding algorithm
  for linear inverse problems,'' \emph{SIAM Journal on Imaging Sciences},
  vol.~2, no.~1, pp. 183--202, 2009.

\bibitem{genzel2017high}
M.~Genzel, ``High-dimensional estimation of structured signals from non-linear
  observations with general convex loss functions,'' \emph{IEEE Transactions on
  Information Theory}, vol.~63, no.~3, pp. 1601--1619, 2017.

\bibitem{dirksen2018robust}
S.~Dirksen and S.~Mendelson, ``Robust one-bit compressed sensing with
  non-gaussian measurements,'' \emph{arXiv preprint arXiv:1805.09409}, 2018.

\bibitem{thrampoulidis2018generalized}
C.~Thrampoulidis and A.~S. Rawat, ``The generalized lasso for sub-gaussian
  measurements with dithered quantization,'' \emph{arXiv preprint
  arXiv:1807.06976}, 2018.

\bibitem{jacques2017time}
L.~Jacques and V.~Cambareri, ``Time for dithering: fast and quantized random
  embeddings via the restricted isometry property,'' \emph{Information and
  Inference: A Journal of the IMA}, vol.~6, no.~4, pp. 441--476, 2017.

\bibitem{xu2018quantized}
C.~Xu and L.~Jacques, ``Quantized compressive sensing with rip matrices: The
  benefit of dithering,'' \emph{arXiv preprint arXiv:1801.05870}, 2018.

\bibitem{yang2016sparse}
Z.~Yang, Z.~Wang, H.~Liu, Y.~Eldar, and T.~Zhang, ``Sparse nonlinear
  regression: Parameter estimation under nonconvexity,'' in \emph{International
  Conference on Machine Learning}, 2016, pp. 2472--2481.

\bibitem{yang2017high}
Z.~Yang, K.~Balasubramanian, and H.~Liu, ``High-dimensional non-gaussian single
  index models via thresholded score function estimation,'' in
  \emph{International Conference on Machine Learning}, 2017, pp. 3851--3860.

\bibitem{yang2017learning}
Z.~Yang, K.~Balasubramanian, Z.~Wang, and H.~Liu, ``Learning non-gaussian
  multi-index model via second-order stein’s method,'' \emph{Advances in
  Neural Information Processing Systems}, 2017.

\bibitem{yang2017stein}
Z.~Yang, K.~Balasubramanian, and H.~Liu, ``On stein's identity and near-optimal
  estimation in high-dimensional index models,'' \emph{arXiv preprint
  arXiv:1709.08795}, 2017.

\bibitem{yap2013stable}
H.~L. Yap, M.~B. Wakin, and C.~J. Rozell, ``Stable manifold embeddings with
  structured random matrices,'' \emph{IEEE Journal on Selected Topics in Signal
  Processing,}, vol.~7, no.~4, pp. 720--730, 2013.

\bibitem{genzel2018mismatch}
M.~Genzel and G.~Kutyniok, ``The mismatch principle: Statistical learning under
  large model uncertainties,'' \emph{arXiv preprint arXiv:1808.06329}, 2018.

\bibitem{talagrand2014gaussian}
M.~Talagrand, ``Gaussian processes and the generic chaining,'' in \emph{Upper
  and Lower Bounds for Stochastic Processes}.\hskip 1em plus 0.5em minus
  0.4em\relax Springer, 2014, pp. 13--73.

\bibitem{oymak2018learning}
S.~Oymak, ``Learning compact neural networks with regularization,''
  \emph{International Conference on Machine Learning}, 2018.

\bibitem{chandrasekaran2012convex}
V.~Chandrasekaran, B.~Recht, P.~A. Parrilo, and A.~S. Willsky, ``The convex
  geometry of linear inverse problems,'' \emph{Foundations of Computational
  mathematics}, vol.~12, no.~6, pp. 805--849, 2012.

\bibitem{McCoy}
D.~Amelunxen, M.~Lotz, M.~B. McCoy, and J.~A. Tropp, ``Living on the edge:
  Phase transitions in convex programs with random data,'' \emph{Inform.
  Inference}, 2014.

\bibitem{tropp2015introduction}
J.~A. Tropp \emph{et~al.}, ``An introduction to matrix concentration
  inequalities,'' \emph{Foundations and Trends{\textregistered} in Machine
  Learning}, vol.~8, no. 1-2, pp. 1--230, 2015.

\bibitem{oymak2018stochastic}
S.~Oymak, ``Stochastic gradient descent learns state equations with nonlinear
  activations,'' \emph{arXiv preprint arXiv:1809.03019}, 2018.

\end{thebibliography}
}

\begin{IEEEbiography}[{\includegraphics[width=1in,height=1.25in,clip,keepaspectratio]{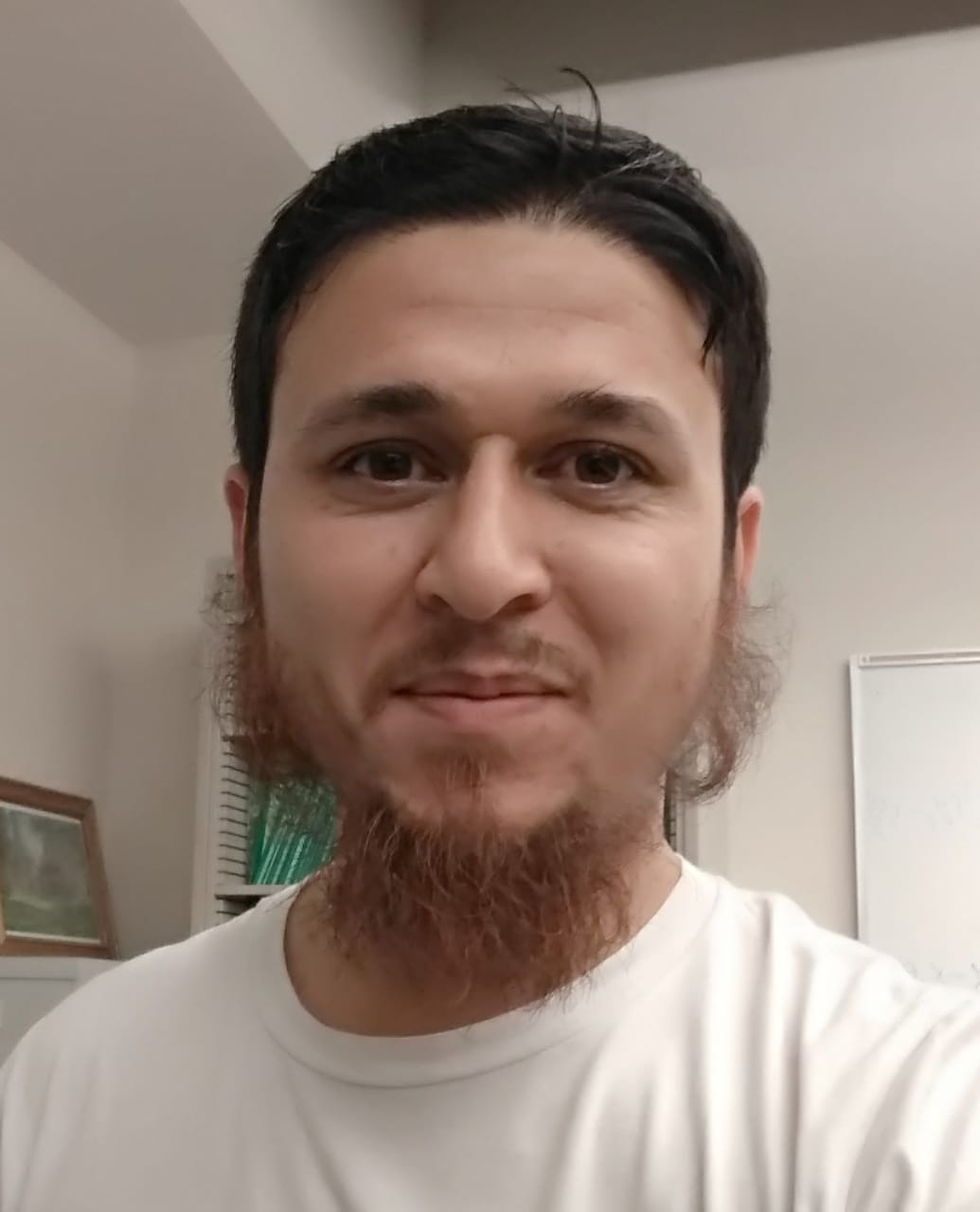}}]{Yahya Sattar}
	received the BS degree in Electrical Engineering from School of Science and Engineering at Lahore University of Management Sciences, Pakistan. He held a position as a Research Assistant at Smart Data, Systems and Applications Lab between 2015 and 2017. He is currently a second year PhD student at the Department of Electrical and Computer Engineering at the University of California, Riverside, CA. His
	research interests include convex/nonconvex optimization, statistical learning theory, signal processing and deep learning.
\end{IEEEbiography}

\begin{IEEEbiography}[{\includegraphics[width=1in,height=1.25in,clip,keepaspectratio]
		{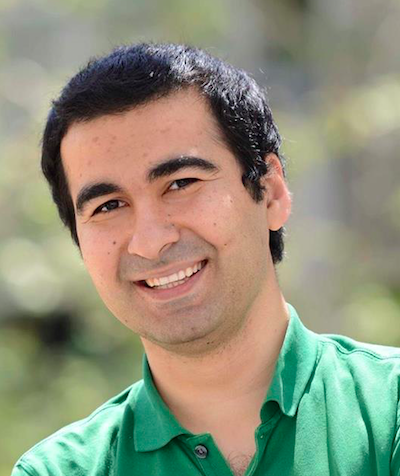}}]{Samet Oymak} is an assistant professor in the Department of Electrical and Computer Engineering, at the University of California, Riverside. He received his MS and PhD degrees from California Institute of Technology; where he was awarded the Wilts Prize for the best thesis in Electrical Engineering. Before joining UCR, he spent time at Google and financial industry, and prior to that he was a fellow at the Simons Institute and a postdoctoral scholar at UC Berkeley. His research explores the mathematical foundations of data science and machine learning by using tools from optimization and statistics. His research interests include mathematical optimization, reinforcement learning, deep learning theory, and high-dimensional problems.
\end{IEEEbiography}

\end{document}